\documentclass[nohyperref]{article}

\usepackage{microtype}
\usepackage{graphicx}
\usepackage{booktabs} 

\usepackage{hyperref}

\usepackage[accepted]{asra_arxiv}  

\usepackage{amsmath}
\usepackage{amssymb}
\usepackage{mathtools}
\usepackage{amsthm}
\usepackage{flushend}
\usepackage{colortbl}
\definecolor{Gray}{gray}{0}
\newcolumntype{P}[1]{>{\centering}p{#1}}
\newcolumntype{M}[1]{>{\centering\arraybackslash}m{#1}}
\usepackage{multirow}
\usepackage{subcaption}
\usepackage[capitalize,noabbrev]{cleveref}

\theoremstyle{plain}

\theoremstyle{definition}

\theoremstyle{remark}

\usepackage[textsize=tiny]{todonotes}

\icmltitlerunning{UnseenNet: Fast Training Detector for Any Unseen Concept}

\begin{document}

\twocolumn[
\icmltitle{UnseenNet: Fast Training Detector for Any Unseen Concept \\ with No Bounding Boxes}



\icmlsetsymbol{equal}{*}

\begin{icmlauthorlist}
\icmlauthor{Asra Aslam}{yyy,sch}
\icmlauthor{Edward Curry}{yyy,sch}
\end{icmlauthorlist}

\icmlaffiliation{yyy}{Insight Centre for the Data Analytics}
\icmlaffiliation{sch}{National University of Ireland Galway, Ireland}

\icmlcorrespondingauthor{Asra Aslam}{asra.aslam.7@gmail.com}

\icmlkeywords{Machine Learning, ICML}

\vskip 0.3in
]



\printAffiliationsAndNotice{}  

\begin{abstract}
Training of object detection models using less data is currently the focus of existing N-shot learning models in computer vision. Such methods use object-level labels and takes hours to train on unseen classes. There are many cases where we have large amount of image-level labels available for training but cannot be utilized by few shot object detection models for training. There is a need for a machine learning framework that can be used for training any unseen class and can become useful in real-time situations. In this paper, we proposed an ``Unseen Class Detector'' that can be trained within a very short time for any possible unseen class without bounding boxes with competitive accuracy. We build our approach on ``Strong'' and ``Weak'' baseline detectors, which we trained on existing object detection and image classification datasets, respectively. Unseen concepts are fine-tuned on the strong baseline detector using only image-level labels and further adapted by transferring the classifier-detector knowledge between baselines. We use semantic as well as visual similarities to identify the source class (i.e. Sheep) for the fine-tuning and adaptation of unseen class (i.e. Goat). Our model (UnseenNet) is trained on the ImageNet classification dataset for unseen classes and tested on an object detection dataset (OpenImages). UnseenNet improves the mean average precision (mAP) by 10\% to 30\% over existing baselines (semi-supervised and few-shot) of object detection on different unseen class splits. Moreover, training time of our model is $<10$ min for each unseen class. Qualitative results demonstrate that UnseenNet is suitable not only for few classes of Pascal VOC but for unseen classes of any dataset or web. Code is available at \url{https://github.com/Asra-Aslam/UnseenNet}.

\end{abstract}

\section{Introduction}
\label{submission}

Detection of objects is one of the significant challenges in computer vision. Multiple object detection models have been proposed to date, including R-CNN \cite{girshick2014rich}, Fast R-CNN \cite{girshick2015fast}, Faster R-CNN \cite{ren2015faster}, SSD \cite{liu2016ssd}, RetinaNet \cite{lin2017focal}, and YOLO \cite{redmon2016you}. The conventional trend is to train such high-performance models on images with bounding box annotations for weeks and detect objects only for certain classes present in object detection datasets (like Pascal VOC \cite{Everingham10}: 20 classes, MCOCO \cite{lin2014microsoft}: 80 classes, OpenImages(OID) \cite{krasin2017openimages}: 600 classes). Recently we started calling such trained classes as ``Seen'' classes, and the remaining classes of the whole world are ``Unseen'' to object detection models. This problem persists not only because we are highly dependent on annotated object bounding boxes based datasets but also we ignore the classification data available on the web, which can be accumulated on request using labels of classes. 
Research focus is mainly on increasing the accuracy by giving few shots \cite{chen2018lstd, kang2019few, yan2019meta, wang2019meta, wu2020multi, wang2020few, sun2021fsce, li2021few}, and where existing image-level labels cannot be utilized. Moreover training time of such approaches is increasing to days while struggling to provide competitive performance. Thus, such approaches are not suitable for real-time applications. Clearly, no settings are available for delivering competitive performance with reduced training time for \textit{unseen} categories.



Several works  \cite{hoffman2014lsda, tang2016large, uijlings2018revisiting, li2018mixed} are bringing a lot of potential in the area of \textit{unseen} concepts by converting classifiers to detectors using image-level labels. These models are trained for a finite number of classes with an adaptation of classifiers into detectors, not for the adaptation to any number of domains. Moreover, they are lagging behind in terms of achieving good accuracy on unseen classes. In this work, we attempt to make detectors using classification data while providing better accuracy and responding to unseen classes in real-time. 

Our work aims to answer the question: Can we train detectors for any possible unseen concept (with only image-level labels) within a limited amount of time while providing competitive accuracy?



We divide the proposed framework ``UnseenNet\footnote{Code will be made publicly available.}'' into two parts; first, we train two detectors ``Strong Baseline'' and ``Weak Baseline'' on 100 classes for weeks ($\sim$300 epochs). Here, we train \textit{strong baseline} on object detection datasets (MCOCO and OID), and \textit{ weak baseline} on image classification dataset (ILSVRC \cite{ILSVRC15}). We use pipeline of YOLO \cite{redmon2018yolov3} with MobileNet \cite{howard2017mobilenets} backbone to train detectors for fast detection and classification. 

The second part is designated for the training of unseen classes. Here, we collect images (with no bounding boxes) from the web using only unseen class names. Then we choose a class (i.e., seen class) from \textit{strong baseline detector} similar to \textit{unseen} class and fine-tune it on collected classification data. We conduct extensive experiments to evaluate the performance of proposed framework (shown in Figure \ref{fig:model}) in low response time for unseen concepts. We show that UnseenNet improves the mean average precision (mAP) by 10 to 30\% over baselines in 10 min of training time, where existing models takes hours (or days) to attain a similar mAP.

\section{Task Definition}
In the case of our training detector for unseen concepts without bounding boxes, we assume that we have access to the object detection datasets (i.e., training images with bounding box annotations for the small number of classes) and image classification datasets (i.e., training images with only image-level labels for the large but finite number of classes). Our objective is to train detectors for any possible unseen concept (i.e., an infinite number of classes) without bounding box annotations within a short time to make them useful for real-time applications. 

\section{Related Work}
\noindent\textbf{Training of Object Detection Models:} Existing object detection models (YOLO \cite{redmon2016you}, RetinaNet \cite{lin2017focal}, SSD \cite{liu2016ssd}, Faster R-CNN \cite{ren2015faster}) require images with bounding box annotations for training, thus falls in the category of fully-supervised models. Such models require training for weeks, even for a limited number of classes, thus cannot process any new/unseen class. We attempt to overcome this limitation by providing training on unseen concepts without bounding boxes in a short training time.


\begin{figure*}
	\begin{center}
		\includegraphics[width=1\textwidth]{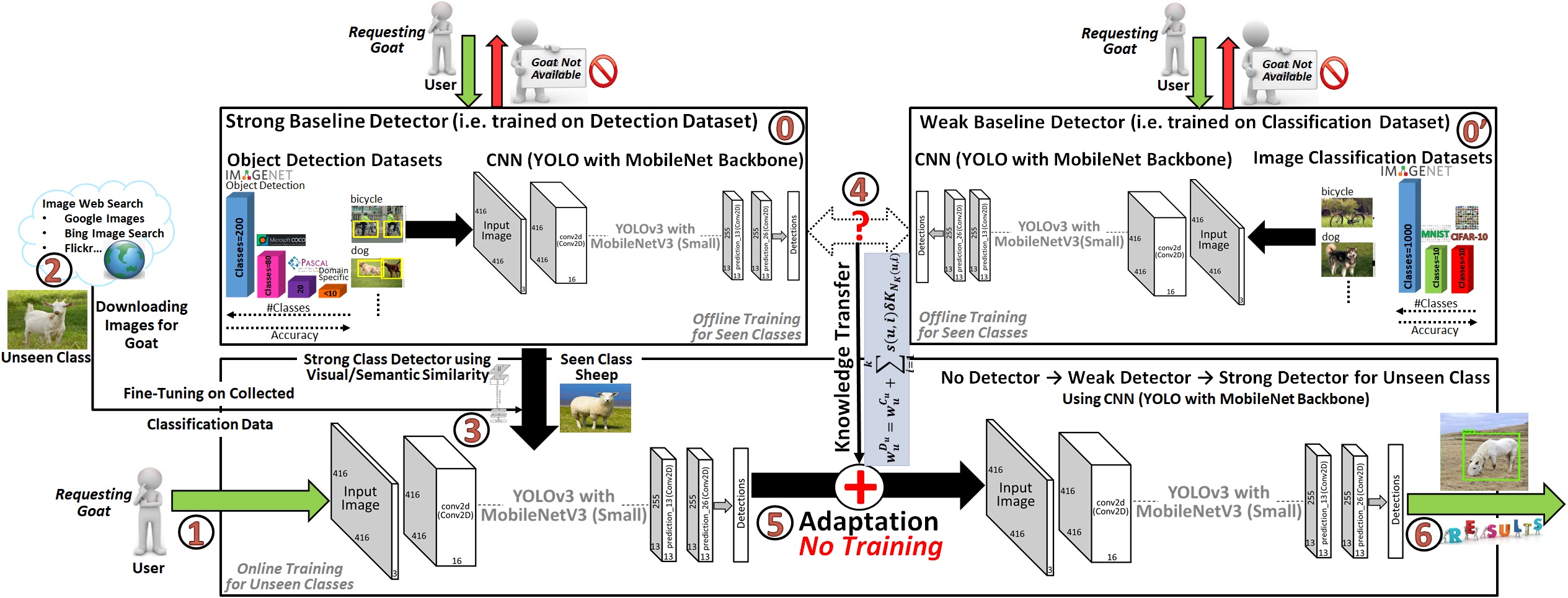}
		\caption{An illustration of our ``UnseenNet'' model. We make use of existing object detection datasets (have bounding box annotations) and image classification datasets (have only image-level labels) and train two separate detectors ``Strong Baseline Detector (0)'' and ``Weak Baseline Detector (0')'' respectively, in advance. (1) On request of any \textit{unseen} concept, (2) we download images using only image-level labels (like goat). (3) \textit{Strong baseline detector} is then fine-tuned on collected images of unseen concepts by labeling the most semantically similar class (like sheep) with the unseen class name (like a goat). (4) At this stage, we also compute the visual similarity of the constructed unseen class detector (trained on classification data) with seen classes of \textit{weak baseline detector}, combine it with semantic similarities, and select top-k classes ranked on comprehensive similarities. (5) Finally, we transfer the knowledge of classifier-detector differences of top classes to the constructed unseen class detector, adapt it into the stronger detector without further training, and (6) return detection results. Note: Baseline detectors (0, 0') are trained offline while other parts (1-6) gets train on unseen class request.}
		\label{fig:model}
	\end{center}
\end{figure*}

\noindent\textbf{Use of Datasets:} Most popular object detection datasets include Pascal VOC \cite{Everingham10}, MCOCO\cite{lin2014microsoft}, OID \cite{krasin2017openimages}, and ILSVRC detection challenge \cite{ILSVRC15}. Though these datasets show promising results on the training of object detection models, indeed fail to fulfill the data requirements of object detection models that could process numerous classes. On the other side, image classification datasets (ImageNet \cite{deng2009imagenet}: 1000, MNIST \cite{lecun2010mnist}: 10, CIFAR-10 \cite{krizhevsky2009learning}: 10 classes) could be useful for training weak detectors regardless of having only image-level labels and iconic images. Presently, the significance of detection/classification datasets only appears in the comparison of deep learning models. We propose settings of utilizing all existing (object-detection, domain-specific, image-classification) datasets to train detectors for unseen classes of real-world scenarios.

\noindent\textbf{Weakly Supervised Detection of Unseen Objects:} Weakly supervised learning \cite{zhu2017soft, oquab2015object, bilen2016weakly, kolesnikov2016improving} is an emerging solution for large-scale unseen concepts. Weakly supervised learning also formulated as a Multiple Instance Learning (MIL) problem. $ WSOD^{2} $\cite{zeng2019wsod2} uses bottom-up object evidences and top-down classification output with an adaptive training mechanism. Multiple MIL based methods appear in literature \cite{tang2017multiple, tang2018weakly, cinbis2016weakly, gokberk2014multi}  for the weakly supervised object localization. Most of these methods use  pre-trained ImageNet \cite{deng2009imagenet}  for initialization. However, our model uses the fastest MobileNetv3 \cite{howard2017mobilenets} as backbone and own trained ``Strong Baseline Detector'' for the initialization. Such methods are evaluated on Pascal VOC \cite{Everingham10}, and Pascal VOC is known in computer vision for a long time; its classes shouldn't be considered unseen.

Another area of related work includes the knowledge transfer from source to target domain. Large Scale Detection through Adaptation (LSDA) \cite{hoffman2014lsda} absed approaches transforms classifiers into object detectors using only image-level labels. Tang et al. \cite{tang2016large} improve LSDA by incorporating informed visual knowledge and semantic similarities. Uijlings et al. \cite{uijlings2018revisiting} proposed a revisit knowledge transfer for detectors training in the weakly supervised settings and outperformed all the baselines. We utilize the concept of transferring knowledge between classifier and detector in our work. Some of the significant advantages of our work includes better accuracy, short training time, flexibility for training any possible unseen (new) concept and no requirement of bounding boxes.

\noindent\textbf{N-Shot Learning:} N-Shot learning is a branch of machine learning which handles the challenge of training a model with only a small amount of data. In terms of terminology, we refer to it as N-way K-Shot-classification, where N is the number of classes and K is the number of labeled training samples from each class. Recently few-shot learning based approaches \cite{chen2018lstd, kang2019few, yan2019meta, wang2019meta, wu2020multi, wang2020few, sun2021fsce, li2021few} are achieving promising results for the task of object detection. Such approaches requires multiple stages of training and thus more training time. Moreover, these approaches does not utilize image-level labels which can be easily acquired rather than object-level labels.
\section{UnseenNet}
We propose an ``UnseenNet'' detector (shown in Figure \ref{fig:model}) which allow user to construct detectors for unseen classes without the need for detection data (no bounding boxes) within the short training time. Our model is based on making use of existing object detection datasets of bounded vocabulary (consists of \textit{seen} concepts) to construct detectors for \textit{unseen} concepts (i.e. unbounded vocabulary) by using the differences between a weak detector (trained on image classification dataset) and a strong detector (trained on object detection datasets). We describe below the construction of our strong and weak baseline detectors offline for \textit{seen} concepts and training of detectors online for \textit{unseen} concepts while investigating the object detection model's training time, which we are referring to as the \textit{response-time} of our model on \textit{unseen} concepts. 

\subsection{Training Baseline Detector Offline for Seen Concept (with Bounded Vocabulary)}
\label{sec:training_offline}
First, we setup an architecture of YOLO with MobileNet backbone and construct two baseline detectors as follows:

\noindent\textbf{Strong Baseline Detector ($D_{S}$)} is a $|K|$ class detector trained on existing object detection datasets. It is a detector that is trained on strong labels (i.e., bounding box annotations). Presently we have taken 100 classes (like LSDA) by considering all classes of MCOCO (80 classes \cite{lin2014microsoft}) and 20 classes of OID \cite{krasin2017openimages}. Please note that 20 classes of Pascal VOC \cite{Everingham10} are also present in MCOCO.

\noindent\textbf{Weak Baseline Detector ($D_{W}$)} is another $|K|$ class detector trained on image classification dataset. We trained it on weak labels (i.e., images-level labels). In this detector, we consider the same classes on which we trained the previous \textit{Strong Baseline Detector}, but we use the ILSVRC \cite{ILSVRC15} classification data. The value of $|K|$ is 100 in both cases.
%

\subsection{Training Online Detector for Unseen Concept (for Unbounded Vocabulary)}
\label{sec:training_online}
On request of an unseen class (u), say \textit{goat}, first our model provides an environment to collect images for `goat'' from the Web using Google Images\footnote{https://github.com/hardikvasa/google-images-download}, Flickr\footnote{https://www.flickr.com/services/api/}, or Bing Image\footnote{https://pypi.org/project/bing-image-downloader/} search. Second, it use the ``Strong Baseline Detector'', and set up a new detector by labeling the most similar \textit{seen} class (like sheep) with \textit{unseen} class (i.e. goat). It is important to note that a similar class (like sheep for goat) can be chosen only using semantic similarity at this stage as visual features of an unseen class cannot be computed before training. Next, we fine-tune the detector on images collected for ``goat''. Now we have a new detector having $|K|$ classes that can detect \textit{goat}. Since \textit{goat} class is trained only on image-level labels, we call it a \textit{weak detector} or simply a classifier ``$C_{u}$'' for unseen class. At this point, our model's response time for \textit{unseen} concepts is equal to the time for fine-tuning. 

We presume that fine-tuning induce a \textit{specific category} bias transformation in the detection network towards class ``goat'' (which is positive from the viewpoint of detecting a class goat). Moreover, this network already encodes a \textit{generic ``background'' category} due to previously trained on detection data (because of strong baseline), which is another positive perspective, as this will automatically make the new detector much more effective in localizing the new class without detection data. Finally, the previous classifier $ C_{u} $ adapts into a corresponding detector $ D_{u} $. This assumes that ``difference between classification and detection of a target object category has a positive correlation with similar categories'' detailed in large scale detection approaches \cite{hoffman2014lsda, tang2016large}. 

Suppose weights of the output layer of $D_{S}$ (Strong Baseline Detector) and $D_{W}$ (Weak Baseline Detector) are $ w^{D_{S}} $ and $ w^{D_{W}} $ respectively. We know that for any \textit{seen} category $i \in K$, final detection weights should be computed as $w_{i}^{D_{S}}=w_{i}^{D_{W}}+\delta_{K_{i}}$, where $\delta_{K_{i}}$ is the difference in weights for the seen category. 

By using this knowledge difference and denoting the $ k^{th} $ nearest neighbor in set $ K $ of category $ u $ as $N_{K}(u,k)$, we adapt the final output detection weights for categories $ u $ as:
\begin{equation}
	w_{u}^{D_{u}}= w_{u}^{C_{u}} + \sum_{i=1}^{k} s(u,i)\delta K_{N_{K}(u,i)}
	\label{main_eq}
\end{equation}
where $k\leq |K|$, and $s(u,i)$ denotes the similarity of seen class (\textit{i}) with unseen class (\textit{u}).


Eq.\ref{main_eq} uses the \textit{weighted nearest neighbor} scheme (\cite{tang2016large, tang2017visual}, where weights are assigned to seen categories based on how similar they are to the unseen category. We select top-k weighted nearest neighbor categories ($s(u,i)$) using Eq.\ref{similarity}. Other than the semantic similarity, we also compute the visual similarity at this stage by using the minimal Euclidean distance between the detection parameters of the last layers of detectors $D_{W}$ and $ C_{u} $. Suppose $ K_{v} $ is the set of visually similar ($ s_{v} $) categories and $ K_{s} $ is the set of semantically similar ($ s_{s} $) categories, then comprehensive similarity $s(u,i)$ for unseen category with seen categories is evaluated as:
\begin{equation}
	s(u,i)=\alpha s_{v}(u,i) + (1-\alpha) s_{s}(u,i),~i \in \{K_{v} \cap K_{s}\}
	\label{similarity}
\end{equation}
where $\alpha \in [0,1]$ is a parameter introduced in literature \cite{tang2016large, tang2017visual} to control the influence of the two similarity measures. We use minimal Euclidean distance between feature distributions of the last layers as visual similarity \cite{hoffman2016adaptive} and naive \textit{path-based} semantic similarity measure of WordNet \cite{pedersen2004wordnet} along with a weighted average scheme to compute the comprehensive similarity ($s(u, i)$) scores. We verify the value $\alpha=0.6$ on simplified similarity measures by analyzing the performance.





Finally, we call this adapted detector ``$D_{u}$'' a \textit{strong detector} for unseen class. We analyze the response-time of our model in Section--\ref{sec:experiments} from the stage of \textit{no detector} to \textit{weak detector ($ C_{u} $)}, and eventually to a \textit{strong detector ($ D_{u} $)}.


\section{Experiments}
\label{sec:experiments}
\subsection{Implementation Details}
\paragraph{Data Preparation} We trained Strong and Weak Baseline Detectors on seen classes offline and performed experiments on unseen classes while having training time constraints


\noindent\textbf{Seen Classes}\\
\indent Strong Baseline Detector Training: In this case, We consider all 80 classes of Microsoft COCO \cite{lin2014microsoft} and 20 classes of Open Images OID \cite{krasin2017openimages} to train a strong baseline detector with bounding box annotations. We select 20 classes from OID by sorting its 600 classes on the basis number of images per class and considering the top 20 with the highest number of images available for training.

\indent Weak Baseline Detector Training: Here, We take the same 100 seen classes, retrieve images with labels from the ISLVRC \cite{ILSVRC15} dataset (i.e., images have no bounding boxes), and train \textit{weak baseline detector} by giving full image size in place of annotations.\\

\noindent\textbf{Unseen Classes}\\
We chose classes from the ILSVRC \cite{ILSVRC15} that are also present in Open Images OID \cite{krasin2017openimages} (consist of 600 classes) and consist of reasonable number of testing images ($ > $100). So that we can evaluate the model on an object detection dataset, which gets trained on image classification dataset. That is, we use the testing dataset of OID for unseen classes to serve as groundtruth in the evaluations.

We also perform qualitative evaluations on additional 16 unseen classes that we downloaded from the web using Google Images API \footnote{https://github.com/hardikvasa/google-images-download}. Such classes are not present in any dataset (Pascal VOC, MCOCO, OID, ImageNet etc.) to-date. This clearly proves our model's significance for unseen concepts (known or unknown).

In our experiments, we consider the pipeline of YOLOv3 \cite{redmon2016you, redmon2018yolov3} and MobileNetv3 \cite{howard2017mobilenets, howard2019searching} for fast detection and classification. Specifically, we used the three layers (38, 117, 165) from the MobileNetv3 (Small) within YOLO to make the prediction\footnote{https://github.com/david8862/keras-YOLOv3-model-set}.


We trained our baseline detectors first on learning rate of $10^{-3}$ till 100 epochs, then we used the decay type exponential till 200 epochs; finally, we used the $10^{-4}$ till 300 epochs as validation loss stopped decreasing near this point. However, for the training of our unseen classes, we used the constant learning rate of $10^{-4}$, which could be increased in future experiments for faster results. We kept the slowest possible learning rate, as our model should serve as the base-work for handling dynamic unseen concepts in short training time. Finally, we utilize the benchmark object detection metrics project \footnote{https://github.com/Adamdad/Object-Detection-Metrics} to evaluate our detections with IOU=$ 0.5 $.


We assume it is essential to specify that ImageNet and Object detection datasets use different name for the same classes, so we are using the vocabulary of WordNet to give a single name to each class and also provide mappings of different datasets with our model.

We used the path vector of WordNet for the semantic similarity measure. Visual similarity is simply computed using the minimal Euclidean distance of weights of the unseen class detector (trained on classification data) and weights of weak baseline detector, which is the same as described in LSDA. Here we use a degree of similarity measure to compute the comprehensive similarity between seen and unseen classes

\begin{figure}
	\begin{center}
		\includegraphics[width=0.25\textwidth]{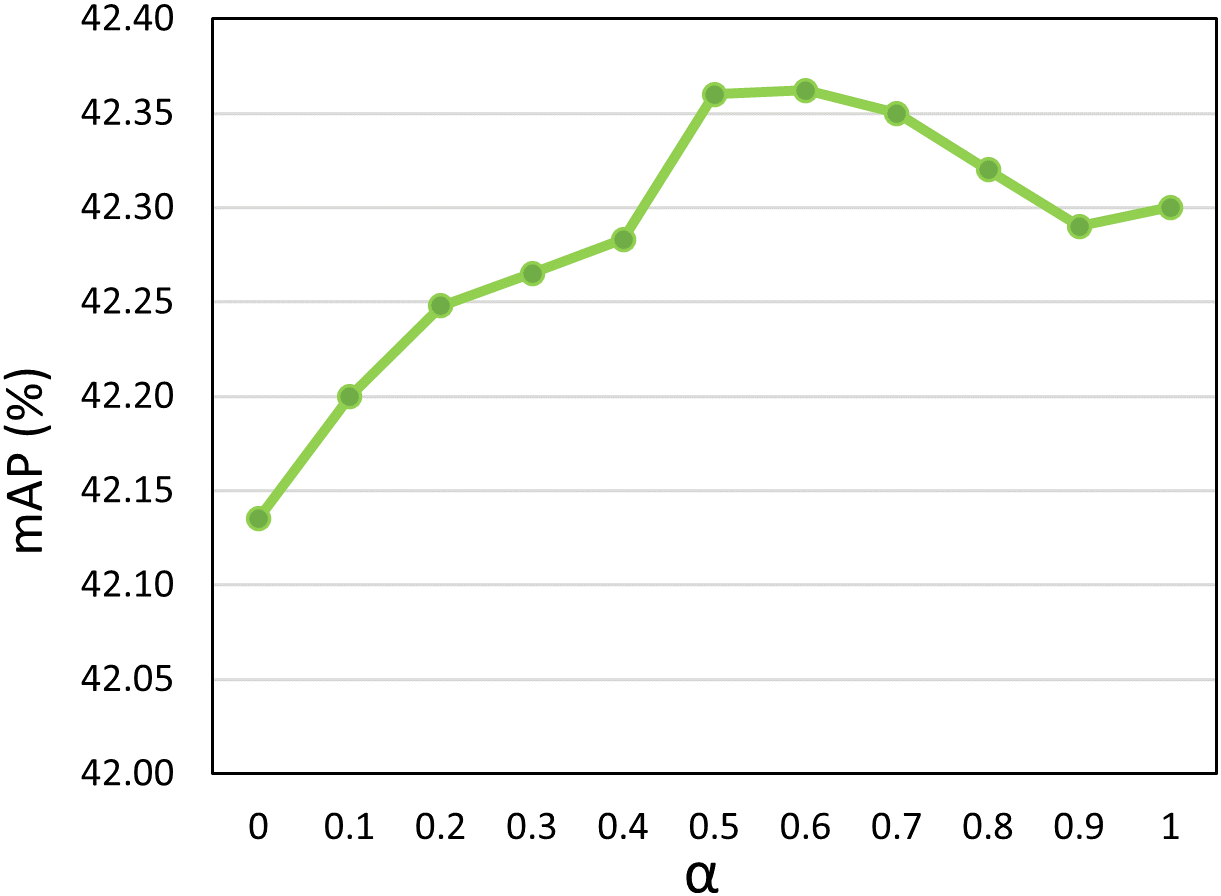}
		\caption{Degree of similarity ($\alpha)$}
		\label{fig:alpha}
	\end{center}
\end{figure}

\noindent\textbf{Degree of Similarity Parameter ($\alpha$)\\}
\label{alpha_details}
To complete the weighted average scheme's evaluations over the simplified (visual and semantic) similarity measures, we also analyzed the value of parameter $ \alpha $, which is responsible for computation of the degree of similarity of the unseen category with seen categories. Figure \ref{fig:alpha} shows the impact of $\alpha$ on mAP, and its possible peak values could be $0.5$, $0.6$, and $0.7$.\\

\noindent\textbf{Estimation of Number of Epochs\\}
We estimated the total number of epochs required to train the model for the designated training time by considering the batch size, number of available training images, and speed of our GPU for the completion of one step. The total number of epochs computed as:
\begin{equation}
	epochs=\frac{Response Time}{((Num~of~Images/Batch~Size)*t)}
\end{equation}
where, ``response time'' denotes the total training time allowed, ``Num of Images'' is the number of available training images, and $ t $ is time GPU takes to complete one step, which is $ 0.465$ sec in our case. Here, ``Num of Images/Batch-Size'' is the number of steps. We used default batch-size $ 16 $. We conducted experiments on NVIDIA TITAN Xp GPU (8 Core Processor$ \times $16), Driver 440.1 with CUDA 10.2.

\begin{table*}
	\centering
	
	\begin{tabular}{ M{1.3cm}  M{2.2cm} M{3.2cm}||M{1.2cm}|M{2.3cm}}
		
		\hline
		\multicolumn{3}{c||}{\textbf{Method}}&\multicolumn{2}{c}{\textbf{``Unseen'' Categories}}\\
		
		&&&\textbf{mAP}&\textbf{Response-Time}\\
		\hline
		\multicolumn{3}{c||}{LSDA \cite{hoffman2014lsda}}&16.33&$ 5.5~hours $\\
		
		\hline
		\multicolumn{3}{c||}{Visual knowledge transfer \cite{tang2016large}}&20.03&$>5.5~hours$\\
		
		\hline
		\multicolumn{3}{c||}{ZSDTR \cite{zheng2021zero}}&20.16&--\\
		\hline
		\multicolumn{3}{c||}{Knowledge Transfer MI \cite{uijlings2018revisiting}}&23.3&--\\
		
		
		\hline
		\multirow{15}{*}{\textbf{UnseenNet}}&\multirow{2}{*}{(Classification Network with No Adapt)}&&22.82&$ 5~min $\\
		&&&27.92&$ 10~min $\\
		&&&27.04&$50~min $\\
		
		\cline{2-5}
		&\multirow{2}{*}{(Class Invariant Adapt \& Specific Class}&&33.36&$ 5~min $\\
		&\multirow{2}{*}{Fine-Tuning)}&&42.03&$ 10~min $\\
		& &&39.77&$50~min $\\
		\cline{2-5}
		
		&&&36.17&$ 5~min $\\
		&\multirow{3}{*}{(Class Invariant}&Weighted Avg NN - 5&\textbf{42.24}&$ \textbf{10~min} $\\
		&\multirow{3}{*}{Adapt,}&&39.79&$50~min $\\
		\cline{3-5}
		
		&\multirow{3}{*}{Specific Class}&&38.55&$ 5~min$\\
		&&Weighted Avg NN - 10&\textbf{42.36}&$ \textbf{10~min} $\\
		&Fine-Tuning,&&39.80&$50~min $\\
		\cline{3-5}
		
		&\& Adapt)&&39.91&$5~min$\\
		&&Weighted Avg NN-100&\textbf{43.07}&$ \textbf{10~min} $\\
		&&&39.88&$50~min $\\
		\hline
	\end{tabular}
	\caption{The mean average precision (mAP) while using ILSVRC for Weak Level labels and MCOCO \& OID for Strong Level labels. First, we show the performance for existing weakly-supervised methods. We also include the performance of semi-supervised LSDA. Row 5--7 shows our model's results on classification network, class invariant adaptation while fine-tuning specific class, then including Classifier to Detector Adaptation. We show the training time (10 min) our model takes to provide similar detection mAP. It is important to note that, inference time of UnseenNet is 9.2fps.}
	\label{full_comparision}
\end{table*}


\subsection{Quantitative Evaluation on Unseen Categories}
\subsubsection{Comparative Analysis with Existing Models}
We compare the performance of the UnseenNet in Table \ref{full_comparision} against weakly supervised object detection models. We show mean average precision (mAP) for unseen categories along with required training time. We evaluate our model by considering different number (5, 10, and 100) of nearest neighbors of ``unseen'' categories with ``seen'' categories while using weighted average nearest neighbor scheme (Eq \ref{main_eq}).

The first 4 rows show the results of existing approaches including LSDA \cite{hoffman2014lsda}, its improved version with visual knowledge transfer \cite{tang2016large}, zero shot learning ZSDTR \cite{zheng2021zero} (as we are also not giving any shots of bounding boxes), and revisiting knowledge transfer MI based approach \cite{uijlings2018revisiting}. We can observe that mean Average Precision (mAP) of existing weakly supervised approaches are low while there training time is very high ($>$5.5 hours) or days and weeks in existing scenarios.  



\begin{figure}
	\begin{center}
		\includegraphics[width=0.49\textwidth]{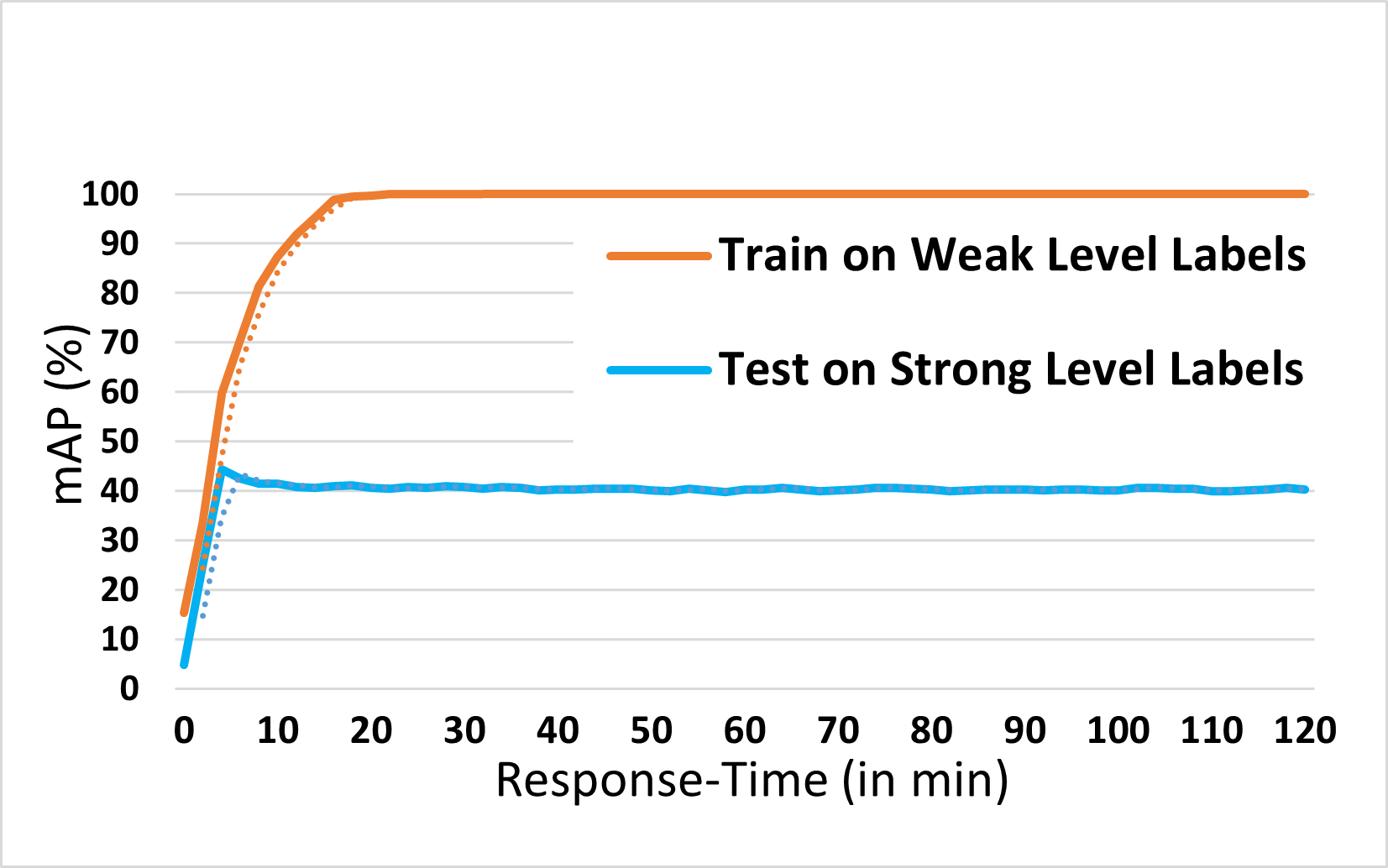}
		\caption{Examples of mAP with Response-Time, For each ``Unseen'' category, we use the top-10 weighted average nearest neighbor ``Seen'' categories for adaptation. This shows that after 10 min of training, mAP does not increase; thus, we choose 20 min as the maximum time. Training for 10 min cross the baseline LSDA, and maximum mAP could be achieved in 5 min.}
		\label{fig:test_train_examples}
	\end{center}
\end{figure}

It is necessary to evaluate our model first by training only on classification data because we are using YOLOv3--MobileNetv3 \cite{redmon2018yolov3, howard2019searching} in contrast to R-CNN--AlexNet \cite{girshick2014rich, krizhevsky2012imagenet}. We show that this amendment improves the performance from $16.33$ to $22.82$. Here we show the mAP for different response time ($5$ min, $10$ min, $50$ min). We choose these response times using testing and training (shown in Figure \ref{fig:test_train_examples}) detail in Section--\ref{sec: Experimental Results with Response-Time}.


Second, we show the mAP using Class Invariant Adapt (Strong Baseline Detector) and fine-tuning the nearest ``seen'' class on target ``unseen'' class classification data. Finally, we apply the specific class adaptation by using the weighted average of ``N'' nearest neighbor classes, where N could be 5, 10, and 100. This step does not require training. We show the final detection performance (average on 100 classes) by indicating our model's total time.
\\
Best results indicate that we can reach from stage of \textbf{no} detector for unseen concepts to weak detector (mAP \textbf{42.03}) and strong detector (mAP \textbf{43.07}) within 10 min of training.

In present case, UnseenNet does not require any shots of bounding box based annotations, and use only image-level labels for finetuning. However, existing few-shot object detection models are showing great promise by providing competitive performance with only few shouts of annotated bounding boxes. Thus, we compared our model performance with recent few-shot detection approaches \cite{chen2018lstd, kang2019few, yan2019meta, wang2019meta, wu2020multi, wang2020few, sun2021fsce, li2021few}. Presently these models perform experiments by considering base classes of Pascal VOC for training and then novel classes also of Pascal VOC; and also in case of Microsoft COCO, base and novel classes belongs to same dataset. However, in our case we used Pascal VOC and Microsoft COCO classes in our strong baseline detector, thus we chose novel classes from OID dataset and still we get better performance than existing few-shot detection methods. Existing approaches used 5-class and 20-class novel splits, so we randomly generated 5-class pairs and 20-class pairs in our dataset of unseen classes. We can observe in Table \ref{few_shot_comparision} for case of 5-class splits we are getting mAP 68.78 which is $>$10\% improvement over existing models. Similarly in case of 20 class splits our performance is mAP 51.09 which is greatest among existing approaches. Again, we trained on classification dataset (ImageNet) and tested on images Open Images dataset which consist of multiple objects in single image. It is important to note that we consider here performance of 10 shots of existing approaches which was best among all shots. Moreover we use only image-level labels thus UnseenNet does not need any shot.

\begin{table}
	\centering
	
	\begin{tabular}{ M{3.8cm} M{1.1cm} M{1.3cm} M{0.5cm}}
		
		\hline
		\multirow{2}{*}{\textbf{Method}}&\multicolumn{2}{c}{\textbf{mAP on Splits}}&\multirow{2}{*}{\textbf{shots}}\\
		&\textbf{5-Class}&\textbf{20-Class}&\\
		\hline
		LSTD \cite{chen2018lstd}&35.27&3.2&\multirow{6}{*}{10}\\
		FSRW \cite{kang2019few}&44.53&5.6&\\
		\multirow{1}{*}{Meta-RCNN \cite{yan2019meta}}&48.33&8.7&\\
		\multirow{1}{*}{MetaDet \cite{wang2019meta}}&44.4&7.1&\\
		MPSR \cite{wu2020multi}&53.1&9.8&\\
		TFA \cite{wang2020few}&48.73&10.0&\\
		FSCE \cite{sun2021fsce}&57.37&11.9&\\
		cos-FSOD \cite{li2021few}&54.77&20.3&\\
		\cline{2-4}
		UnseenNet&\textbf{68.78}&\textbf{51.29}&0\\
		\hline

		\hline
	\end{tabular}
	\caption{Comparison with Few-Shot Detection methods}
	\label{few_shot_comparision}
\end{table}
\subsubsection{Experimental Results with Response-Time}
\label{sec: Experimental Results with Response-Time}
To retrieve the range of response-time effective in our model, we train each category until the point testing accuracy starts to decrease (to avoid over-fitting). We show average performance of all unseen concepts with training time in Figure \ref{fig:test_train_examples}. Please note here we compute the total number of epochs for varying the training time (detail in Section--\ref{alpha_details}). We first train our model on weak level labels (i.e., without bounding boxes) and then test on strong labels (i.e., with bounding boxes). Here weak labels are taken from ImageNet classification data and strong labels are taken from OID dataset. We observe that the maximum mAP of each class could be achieved within $10$ min of training. After that, mAP decreases and remain constant. However, we recommend $10$ min of training to attain maximum mAP $43.07$ to avoid any unexpected reduction in mAP due to over-fitting.

\subsubsection{Experimental Results with Unseen Concepts}
We present an analysis mAP with similarities of unseen categories with \textit{seen} categories (top-10) for few examples of our unseen classes. The simple average similarity score:
\begin{equation}
	s_{j}=\frac{\sum_{i=1}^{m} s(j,i)}{m}
\end{equation}
where m is $ 10 $ presently and $s(j,i)$ is the comprehensive similarity (shown in Eq. \ref{similarity}) between unseen ($ j $) and seen ($ i $) category computed using $\alpha$=0.6. It shows if we have \textit{unseen} classes (like building, pasta, salad)  more similar to \textit{seen} classes, then our model have high probability of giving high performance with the exception for small size objects or availability of less training data.
%
\begin{figure}
	\begin{center}
		\includegraphics[width=0.48\textwidth]{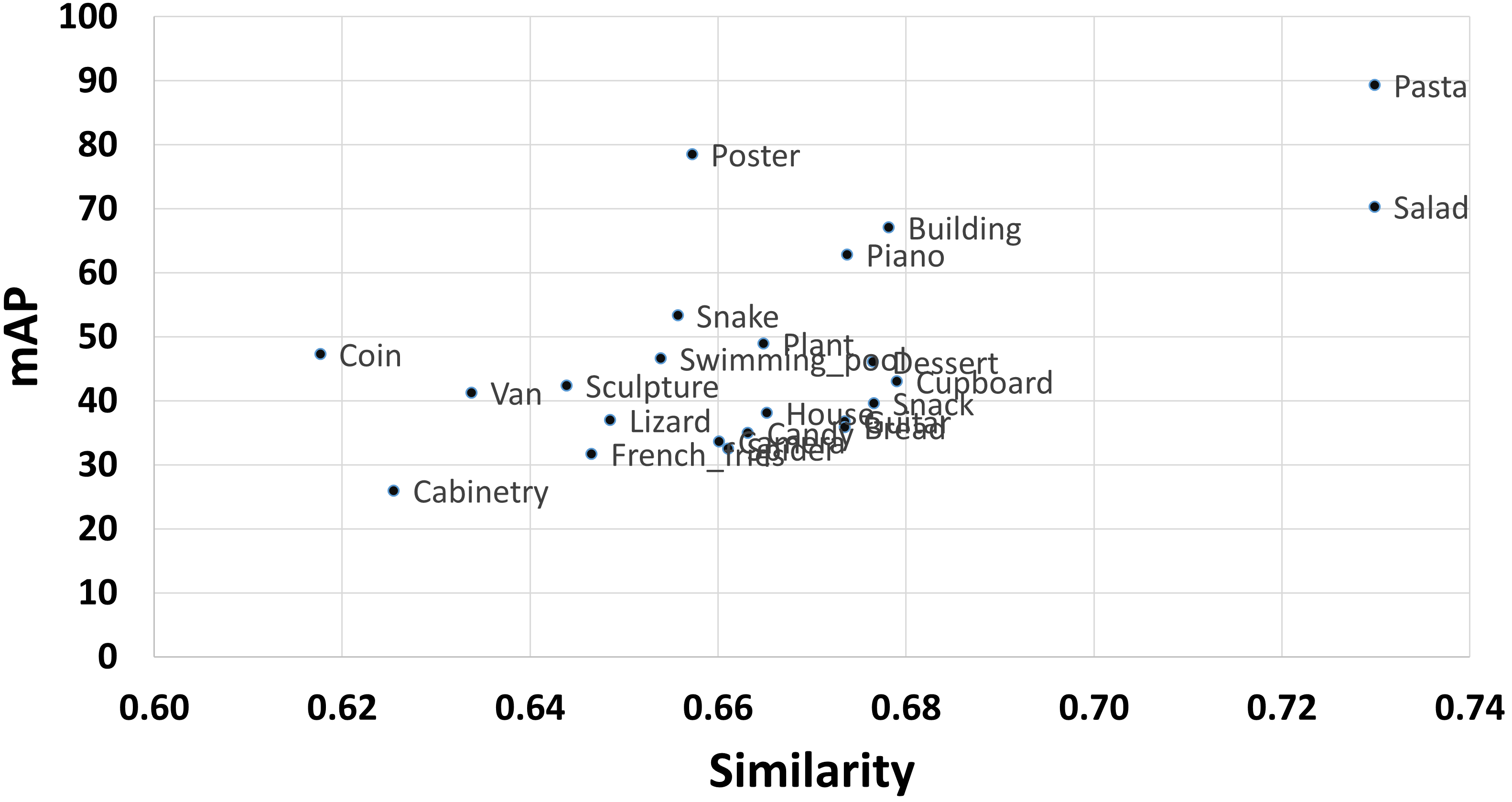}
		\caption{mAP of unseen classes with similarity scores}
		\label{fig:similarities}
	\end{center}
\end{figure}
\begin{figure*}	
	\begin{subfigure}{.12\textwidth}
		\begin{center}
			\includegraphics[width=2cm,height=1.7cm]{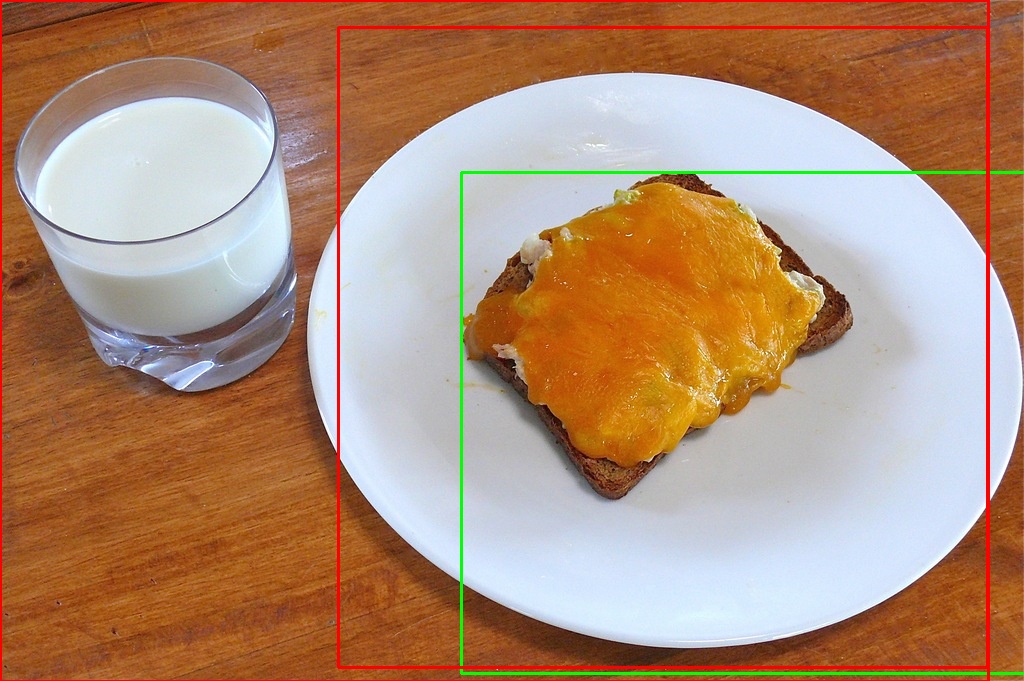}
			\caption{Bread}
			\label{fig:accuracy_time_scratch_yolo}
		\end{center}	
	\end{subfigure}%
	\hfill
	\begin{subfigure}{.12\textwidth}
		\begin{center}
			\includegraphics[width=2.1cm,height=1.7cm]{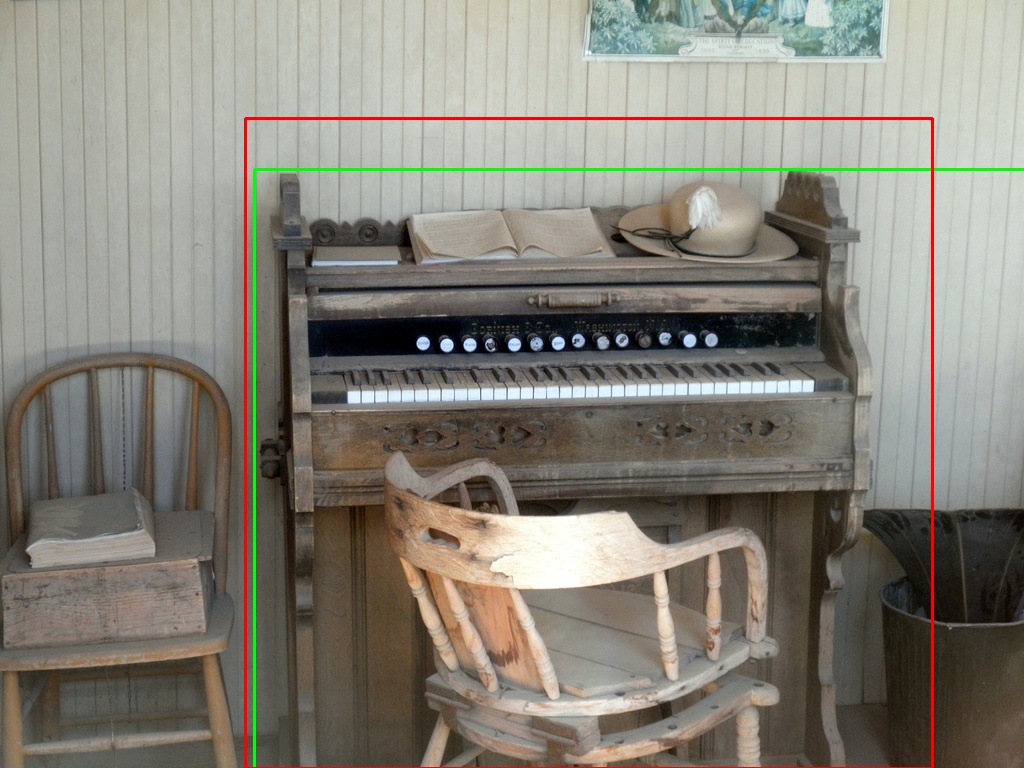} 
			\caption{Piano}
			\label{fig:accuracy_time_scratch_ssd}
		\end{center}
	\end{subfigure}
	\hfill
	\begin{subfigure}{.12\textwidth}
		\begin{center}
			\includegraphics[width=2.15cm,height=1.7cm]{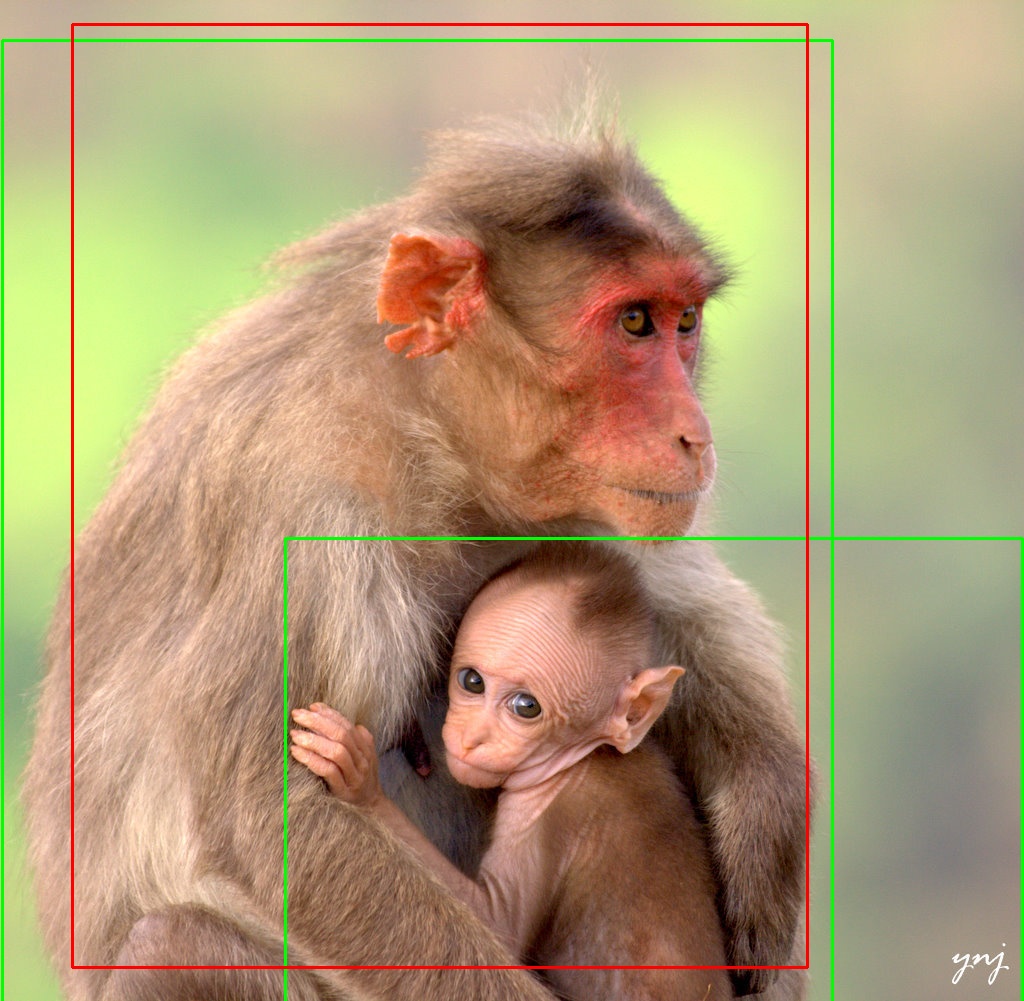} 
			\caption{Monkey}
			\label{fig:accuracy_time_scratch_retinanet}
		\end{center}
	\end{subfigure}	
	\hfill
	\begin{subfigure}{.12\textwidth}
		\begin{center}
			\includegraphics[width=2.15cm,height=1.7cm]{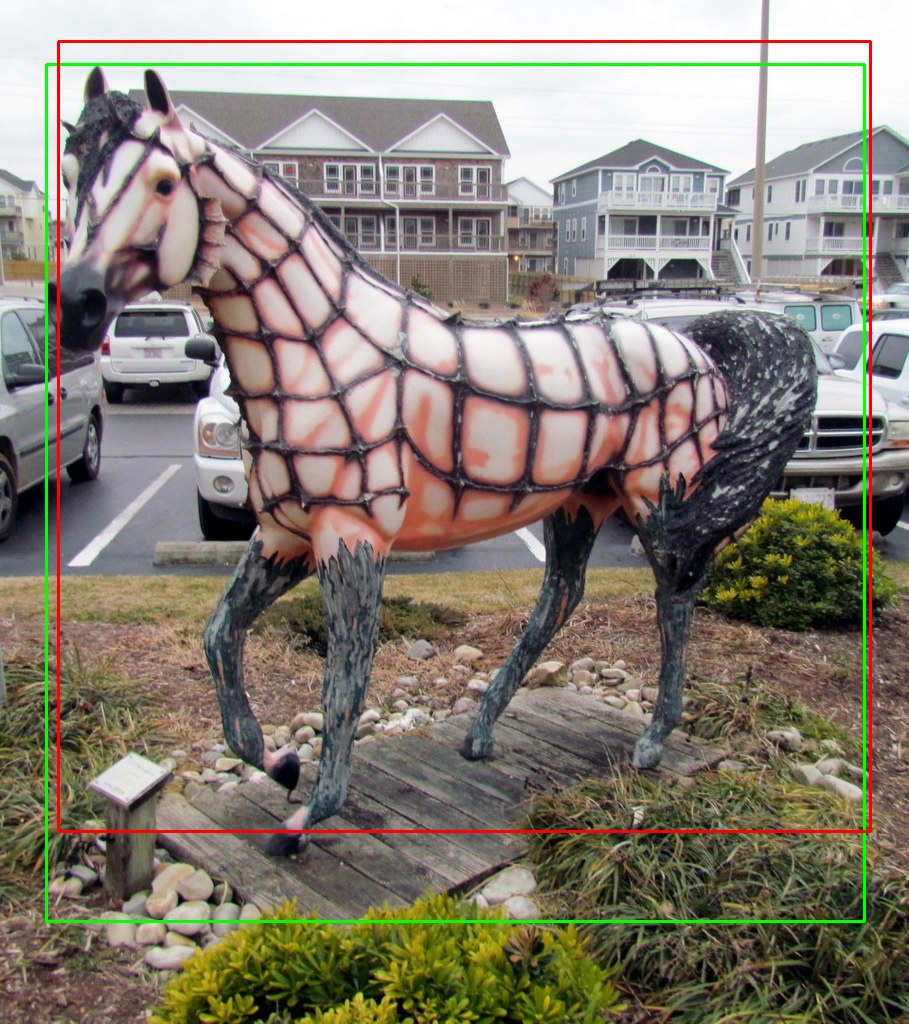} 
			\caption{Sculpture}
			\label{fig:accuracy_time_scratch_retinanet}
		\end{center}
	\end{subfigure}	
	\hfill
	\begin{subfigure}{.12\textwidth}
		\begin{center}
			\includegraphics[width=2.1cm,height=1.7cm]{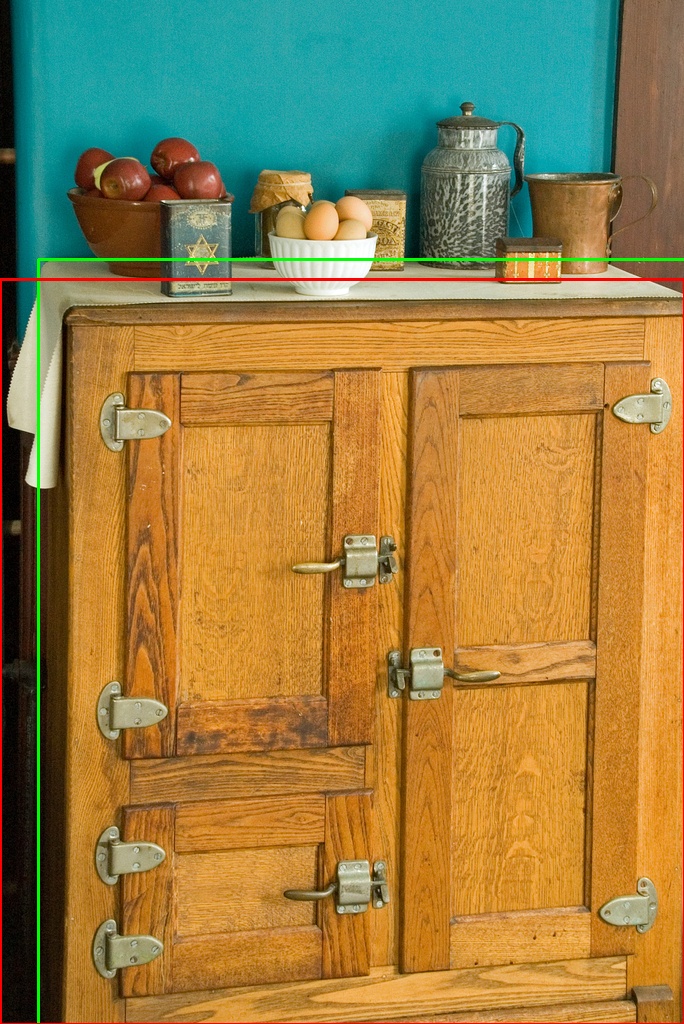} 
			\caption{Cupboard}
			\label{fig:accuracy_time_scratch_retinanet}
		\end{center}
	\end{subfigure}	
	\hfill
	\begin{subfigure}{.12\textwidth}
		\begin{center}
			\includegraphics[width=2.15cm,height=1.7cm]{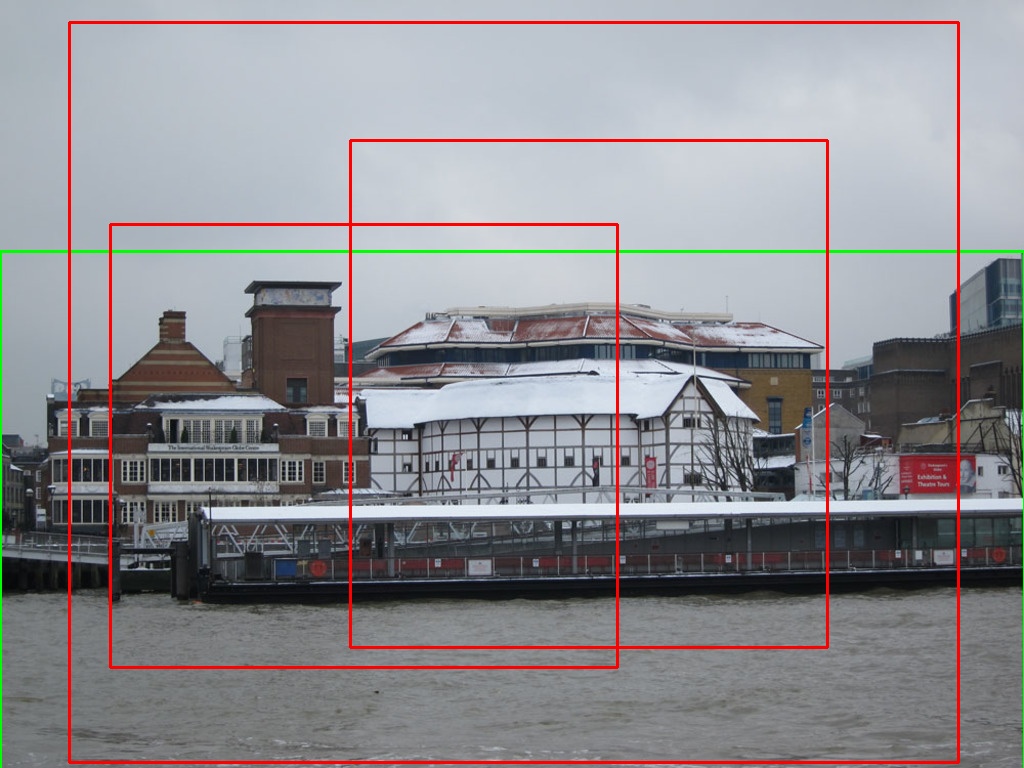} 
			\caption{Building}
			\label{fig:accuracy_time_scratch_retinanet}
		\end{center}
	\end{subfigure}	
	\hfill
	\begin{subfigure}{.12\textwidth}
		\begin{center}
			\includegraphics[width=2.2cm,height=1.7cm]{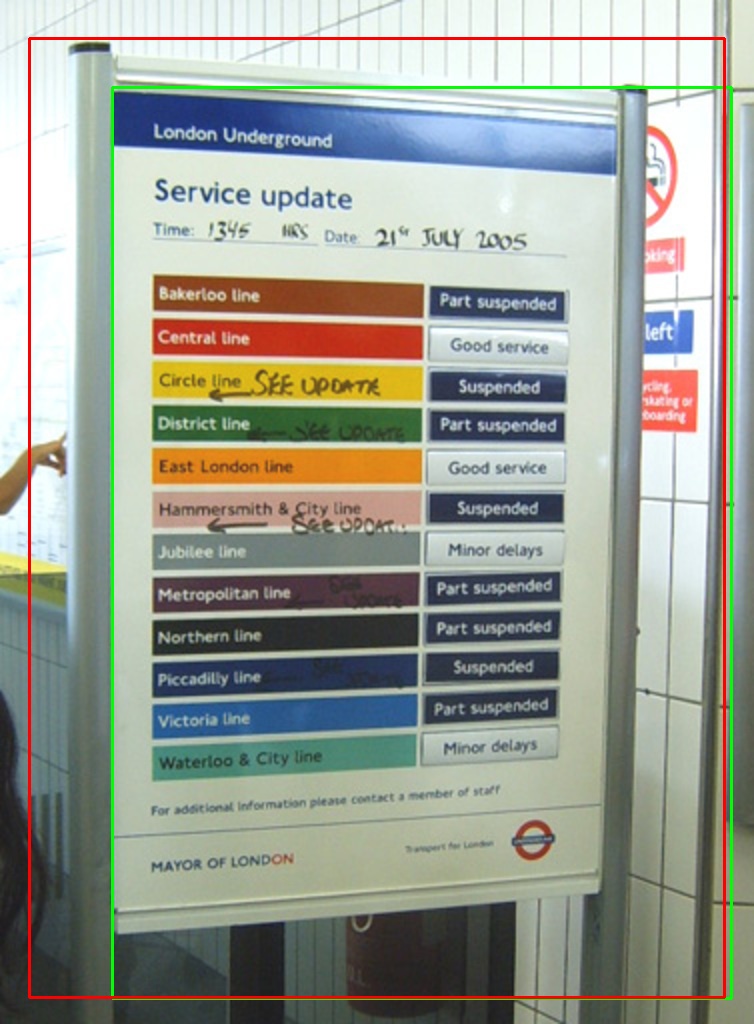} 
			\caption{Poster}
			\label{fig:accuracy_time_scratch_retinanet}
		\end{center}
	\end{subfigure}	
	\hfill
	\begin{subfigure}{.12\textwidth}
		\begin{center}
			\includegraphics[width=2.05cm,height=1.7cm]{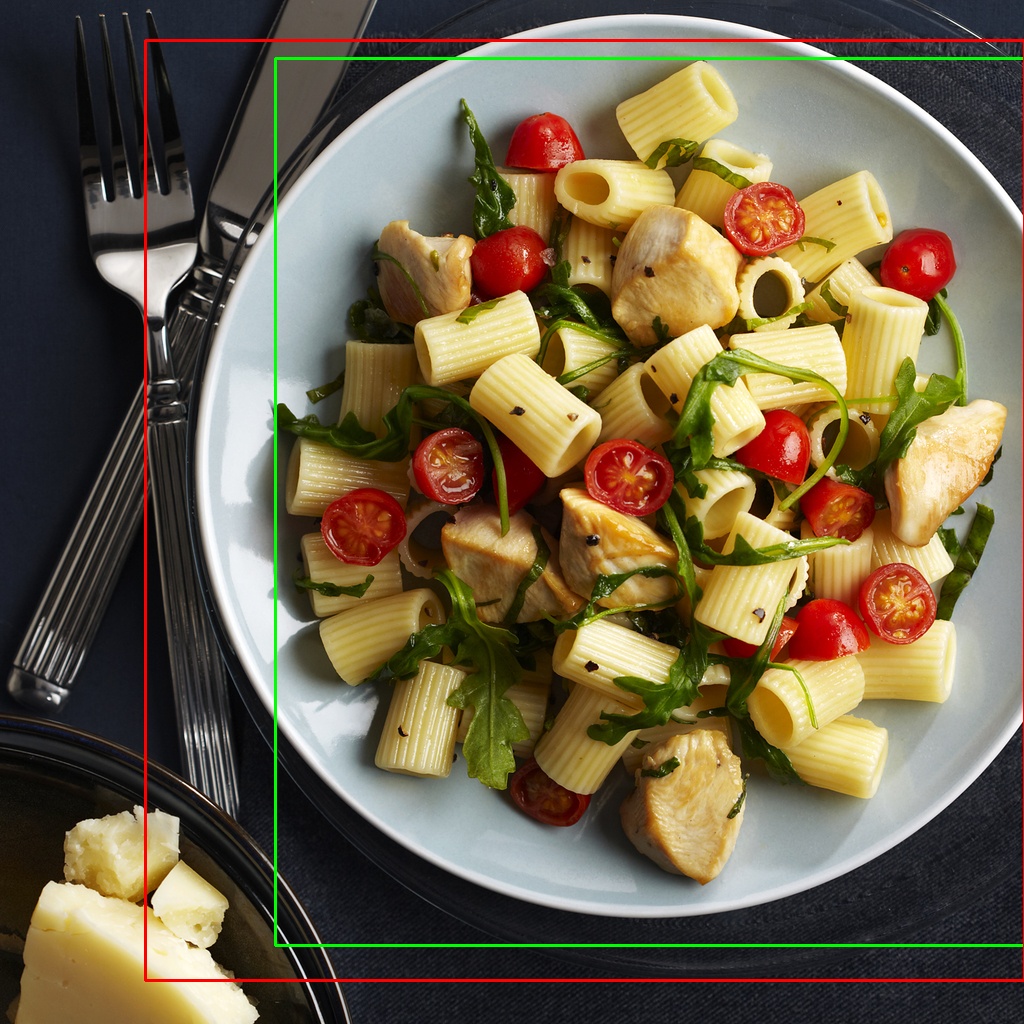} 
			\caption{Pasta}
			\label{fig:accuracy_time_scratch_retinanet}
		\end{center}
	\end{subfigure}	
	
	\begin{subfigure}{.12\textwidth}
		\begin{center}
			\includegraphics[width=2cm,height=1.7cm]{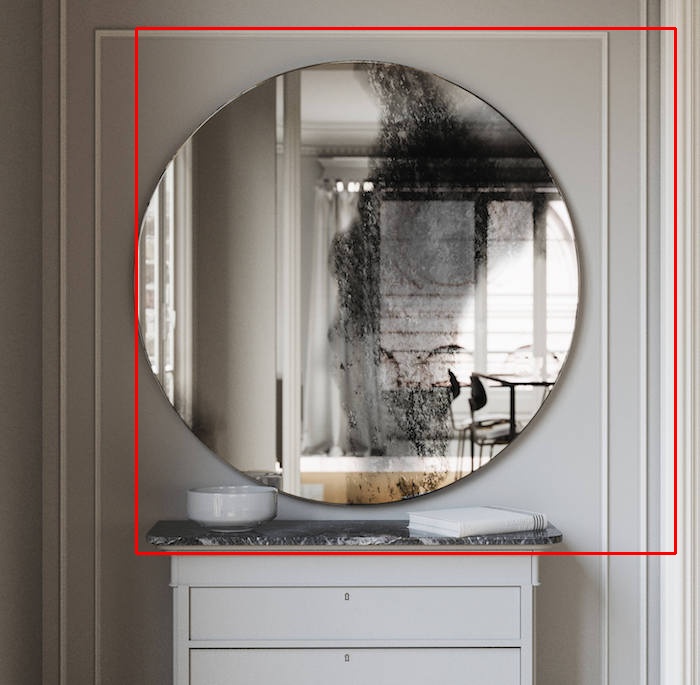} 
			\caption{Mirror}
			\label{fig:accuracy_time_scratch_yolo}
		\end{center}	
	\end{subfigure}%
	\hfill
	\begin{subfigure}{.12\textwidth}
		\begin{center}
			\includegraphics[width=2cm,height=1.7cm]{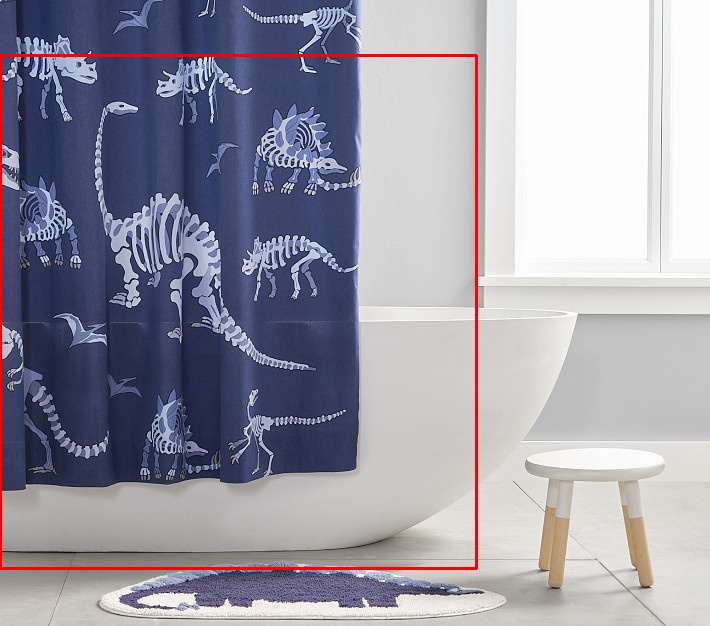} 
			\caption{Curtain}
			\label{fig:accuracy_time_scratch_ssd}
		\end{center}
	\end{subfigure}
	\hfill
	\begin{subfigure}{.136\textwidth}
		\begin{center}
			\includegraphics[width=2cm,height=1.7cm]{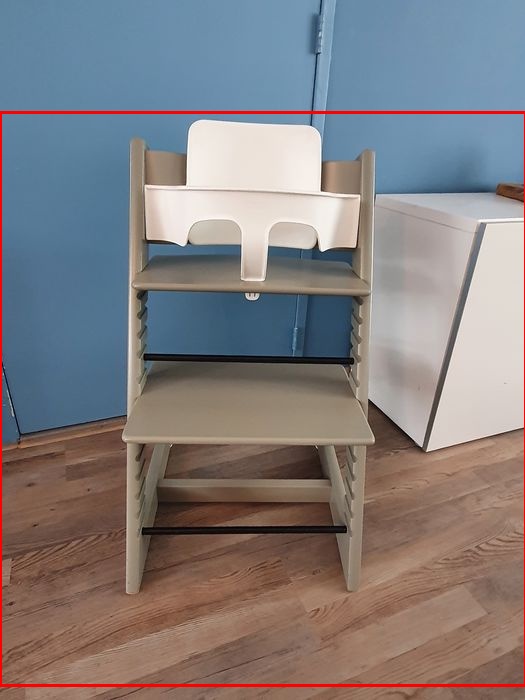} 
			{\tiny \caption{BabyHighchair}}
			\label{fig:accuracy_time_scratch_retinanet}
		\end{center}
	\end{subfigure}	
	\hfill
	\begin{subfigure}{.116\textwidth}
		\begin{center}
			\includegraphics[width=2cm,height=1.7cm]{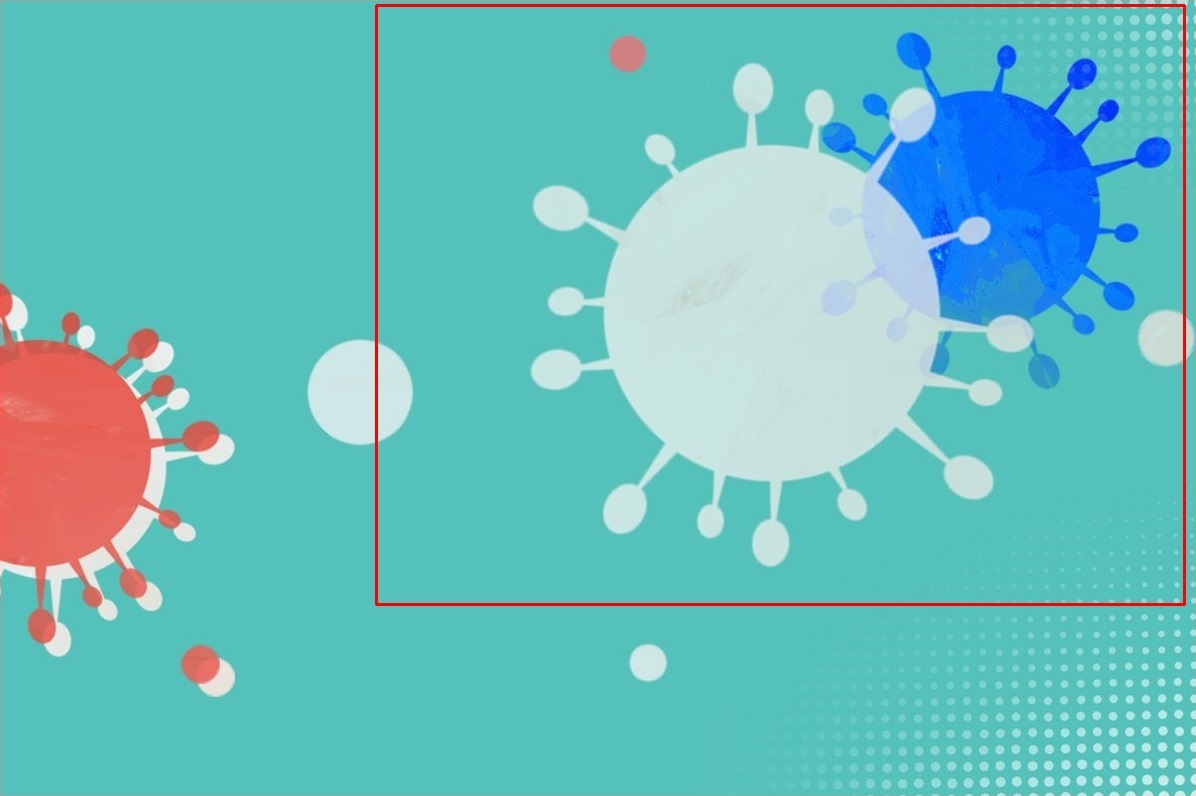} 
			\caption{COVID Icon}
			\label{fig:accuracy_time_scratch_retinanet}
		\end{center}
	\end{subfigure}	
	\hfill
	\begin{subfigure}{.12\textwidth}
		\begin{center}
			\includegraphics[width=2cm,height=1.7cm]{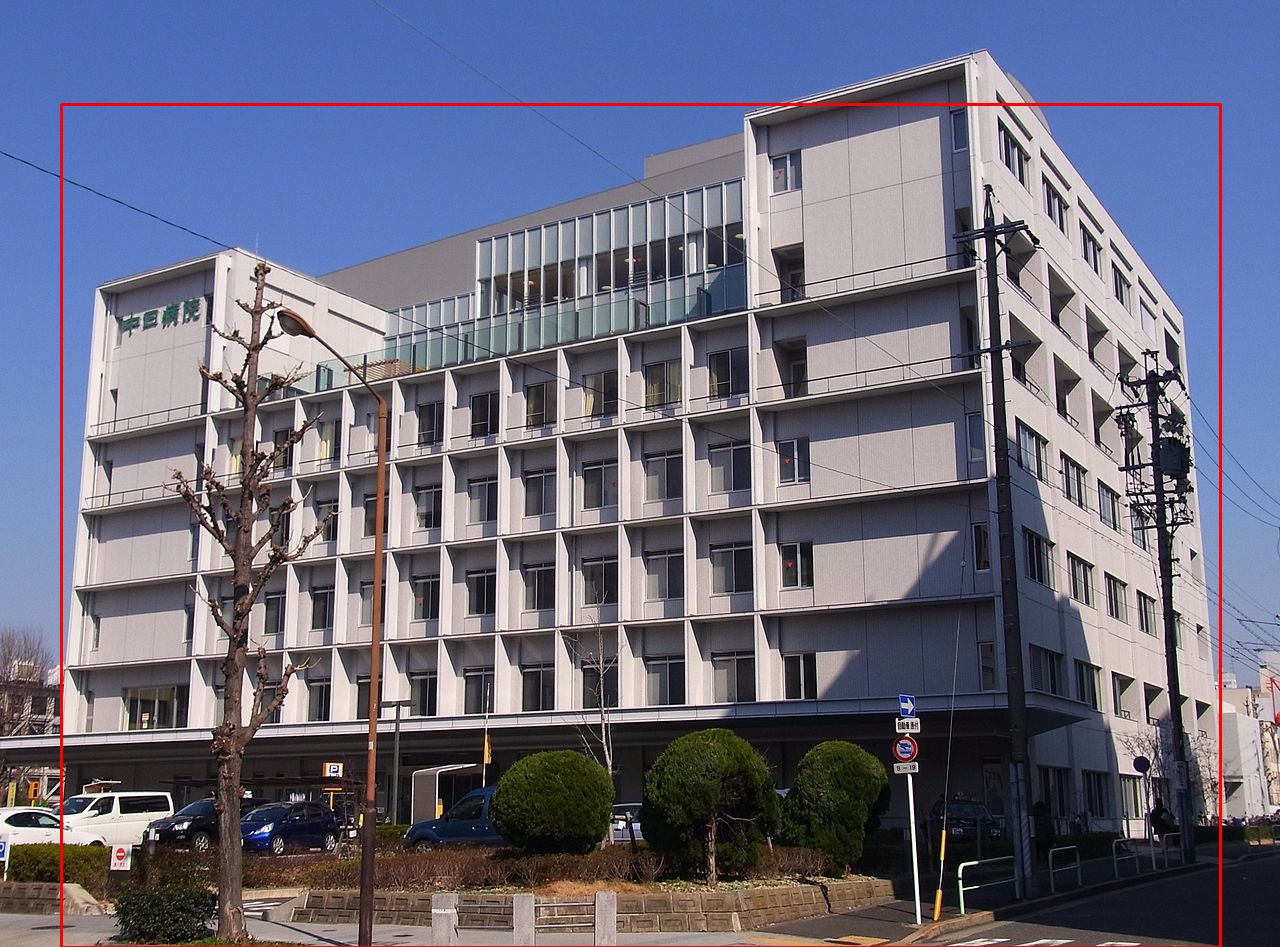} 
			\caption{Hospital}
			\label{fig:accuracy_time_scratch_retinanet}
		\end{center}
	\end{subfigure}	
	\hfill
	\begin{subfigure}{.11\textwidth}
		\begin{center}
			\includegraphics[width=2cm,height=1.7cm]{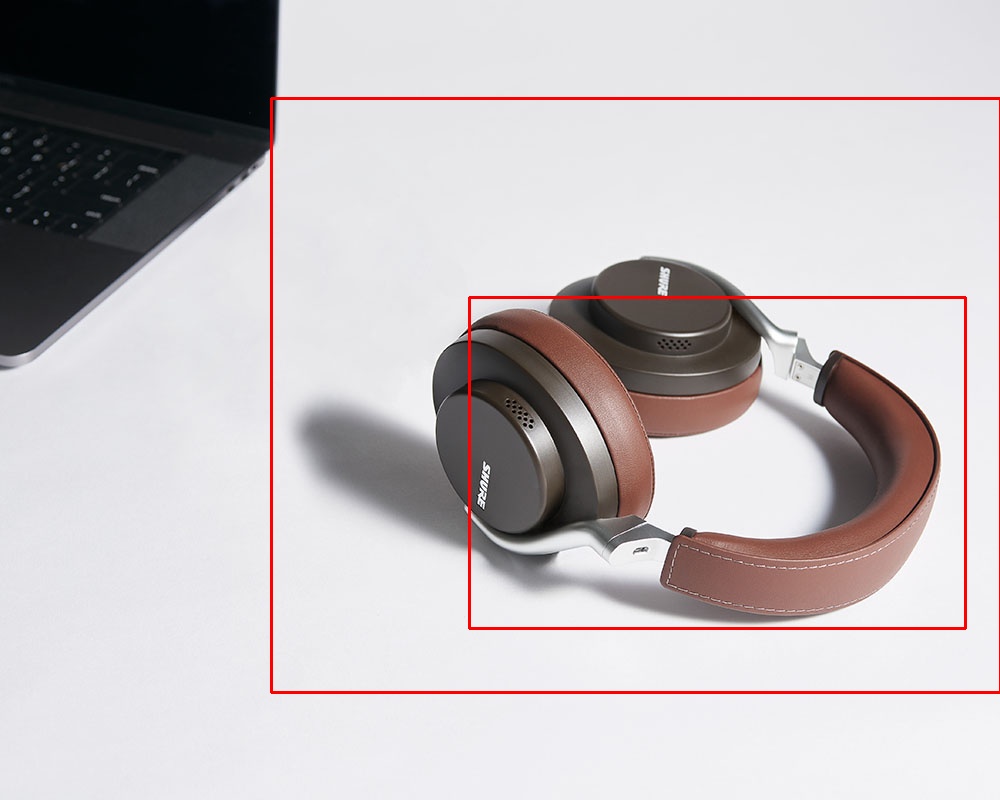} 
			\caption{Headphone}
			\label{fig:accuracy_time_scratch_retinanet}
		\end{center}
	\end{subfigure}	
	\hfill
	\begin{subfigure}{.11\textwidth}
		\begin{center}
			\includegraphics[width=2cm,height=1.77cm]{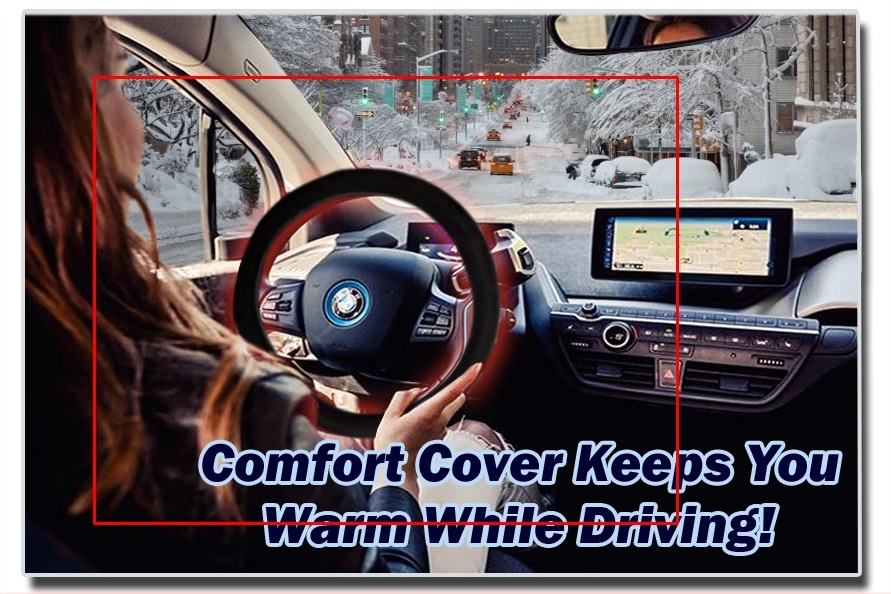} 
			\caption{Steering}
			\label{fig:accuracy_time_scratch_retinanet}
		\end{center}
	\end{subfigure}	
	\hfill
	\begin{subfigure}{.11\textwidth}
		\begin{center}
			\includegraphics[width=2.05cm,height=1.7cm]{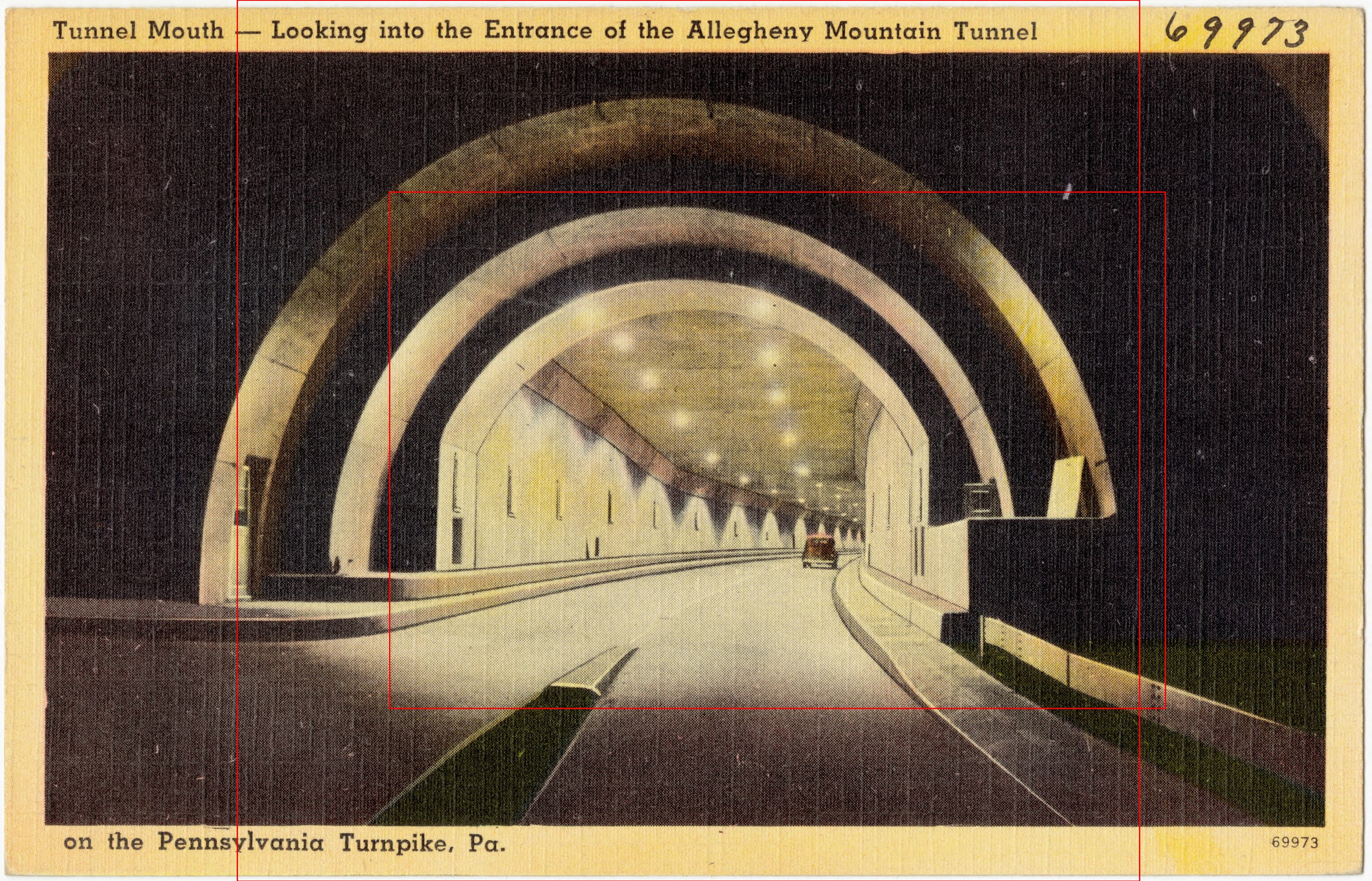} 
			\caption{Tunnel}
			\label{fig:accuracy_time_scratch_retinanet}
		\end{center}
	\end{subfigure}

	\caption{Examples of correct detections. 2nd row unseen classes are downloaded online, no groundtruth available to date.}
	\label{fig:example_correct_det}
\end{figure*}
\begin{figure*}	
	\begin{subfigure}{.12\textwidth}
		\begin{center}
			\includegraphics[width=2cm,height=1.7cm]{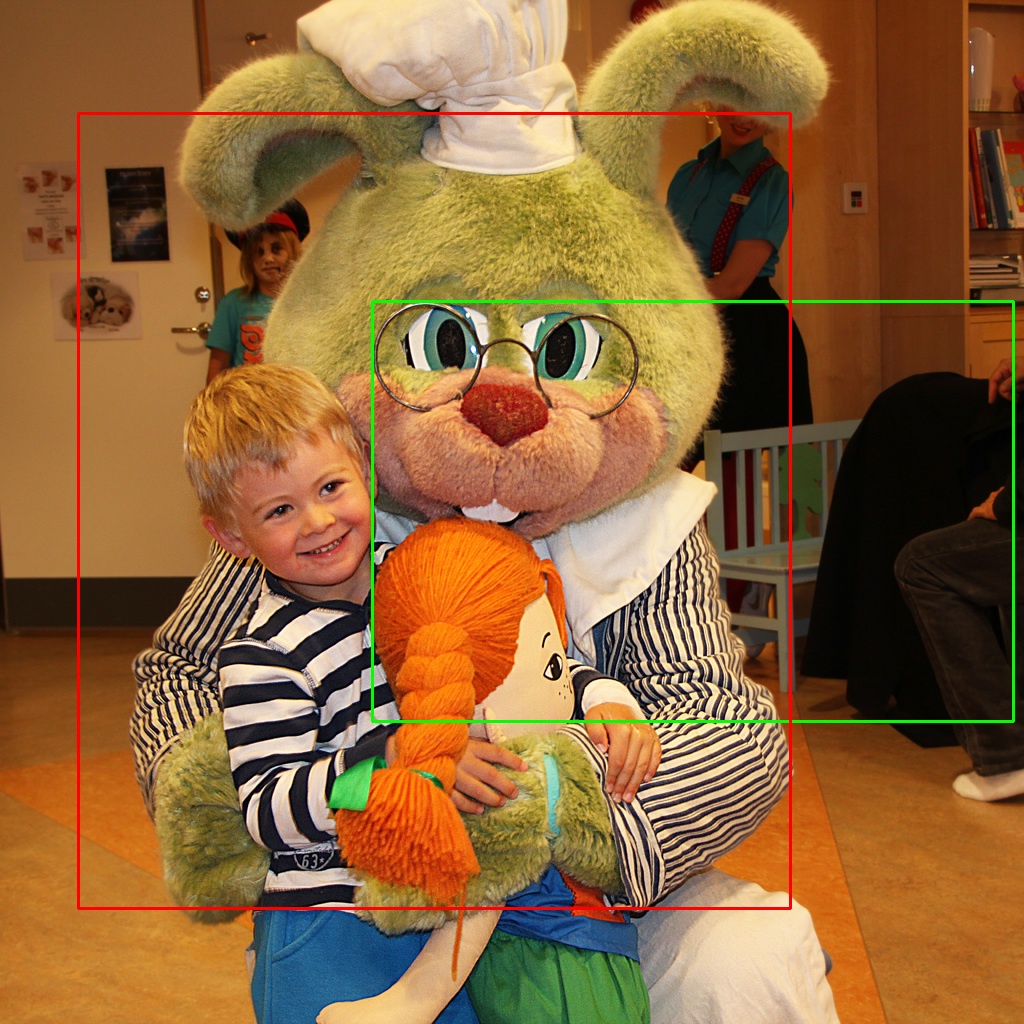} 
			\caption{Glasses}
			\label{fig:accuracy_time_scratch_yolo}
		\end{center}	
	\end{subfigure}%
	\hfill
	\begin{subfigure}{.12\textwidth}
		\begin{center}
			\includegraphics[width=2.15cm,height=1.7cm]{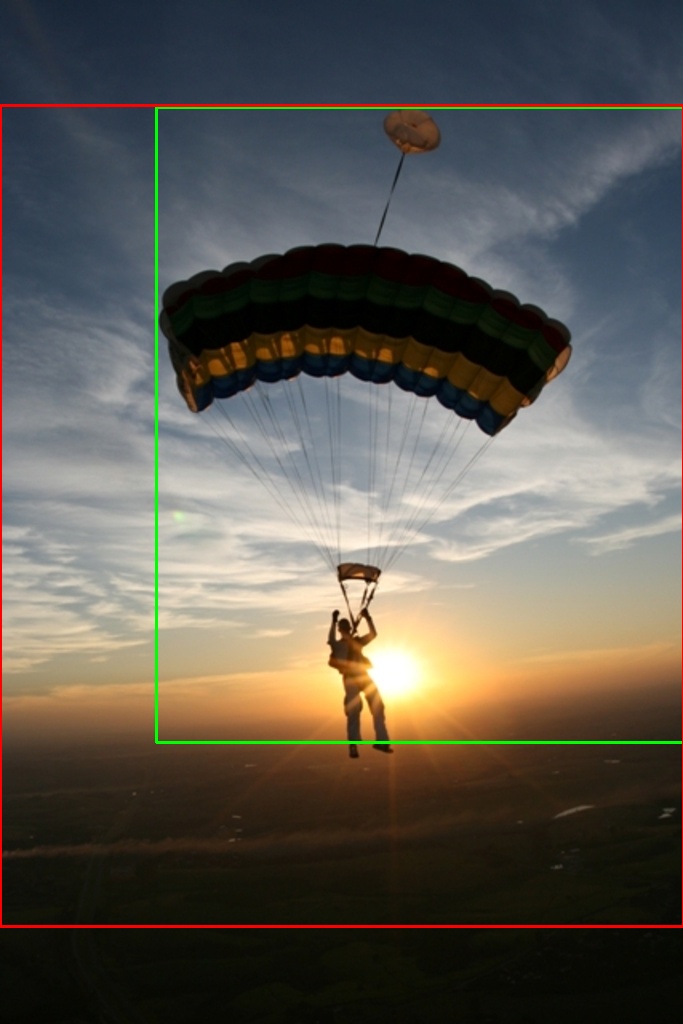} 
			\caption{Parachute}
			\label{fig:accuracy_time_scratch_ssd}
		\end{center}
	\end{subfigure}
	\hfill
	\begin{subfigure}{.12\textwidth}
		\begin{center}
			\includegraphics[width=2.15cm,height=1.7cm]{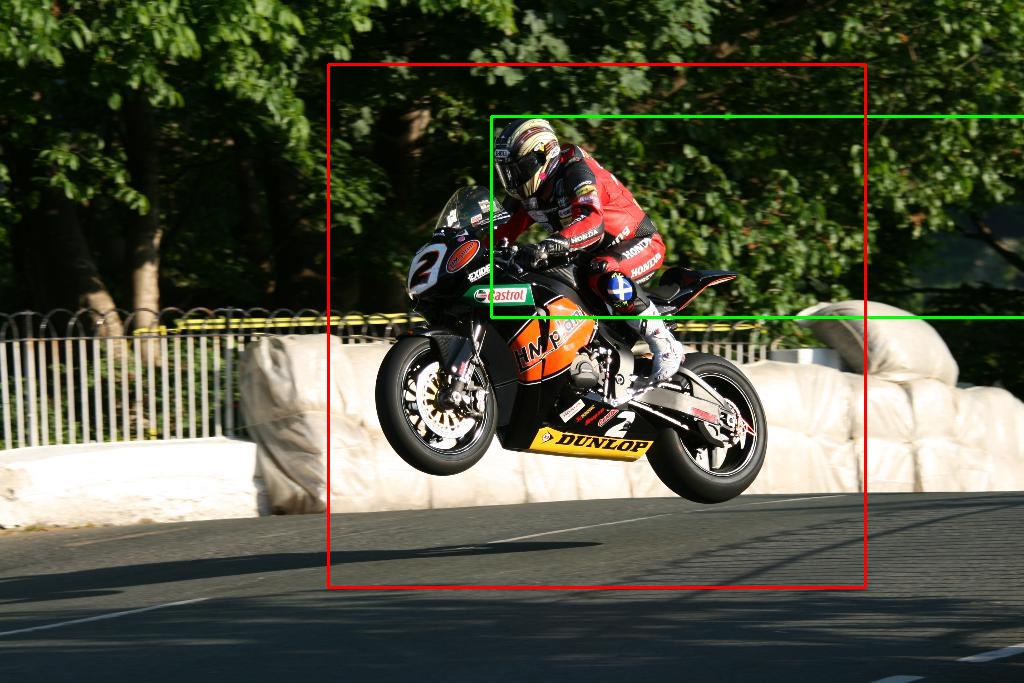} 
			\caption{Helmet}
			\label{fig:accuracy_time_scratch_retinanet}
		\end{center}
	\end{subfigure}	
	\hfill
	\begin{subfigure}{.12\textwidth}
		\begin{center}
			\includegraphics[width=2.15cm,height=1.7cm]{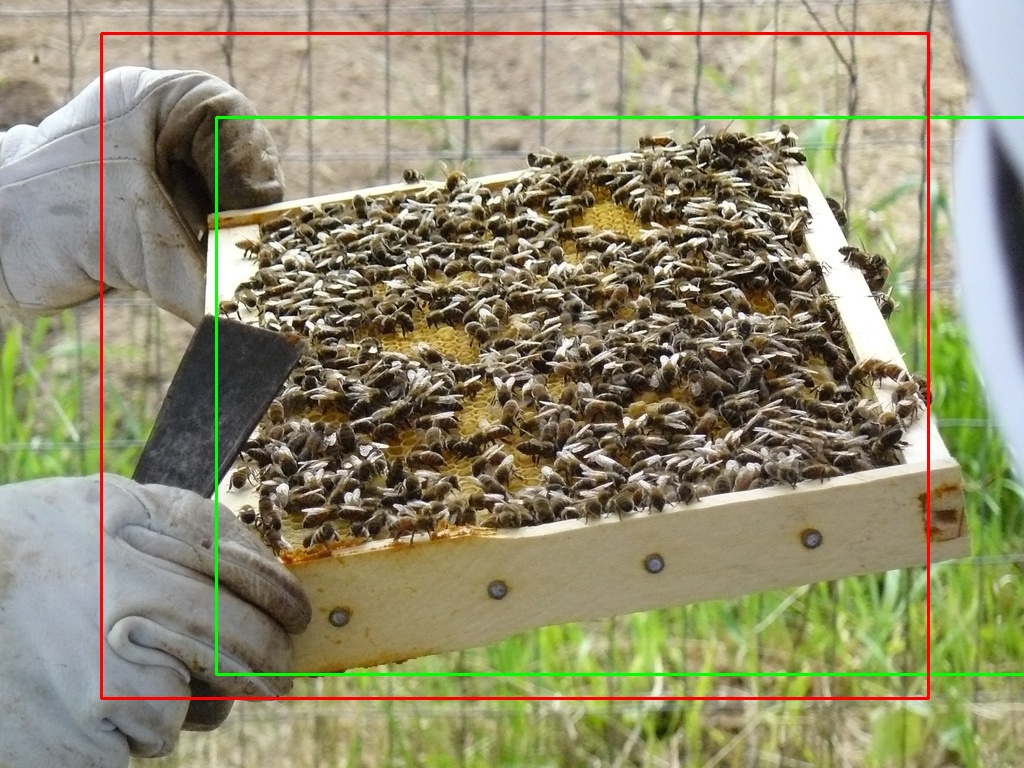} 
			\caption{Bee}
			\label{fig:accuracy_time_scratch_retinanet}
		\end{center}
	\end{subfigure}	
	\hfill
	\begin{subfigure}{.12\textwidth}
		\begin{center}
			\includegraphics[width=2.13cm,height=1.7cm]{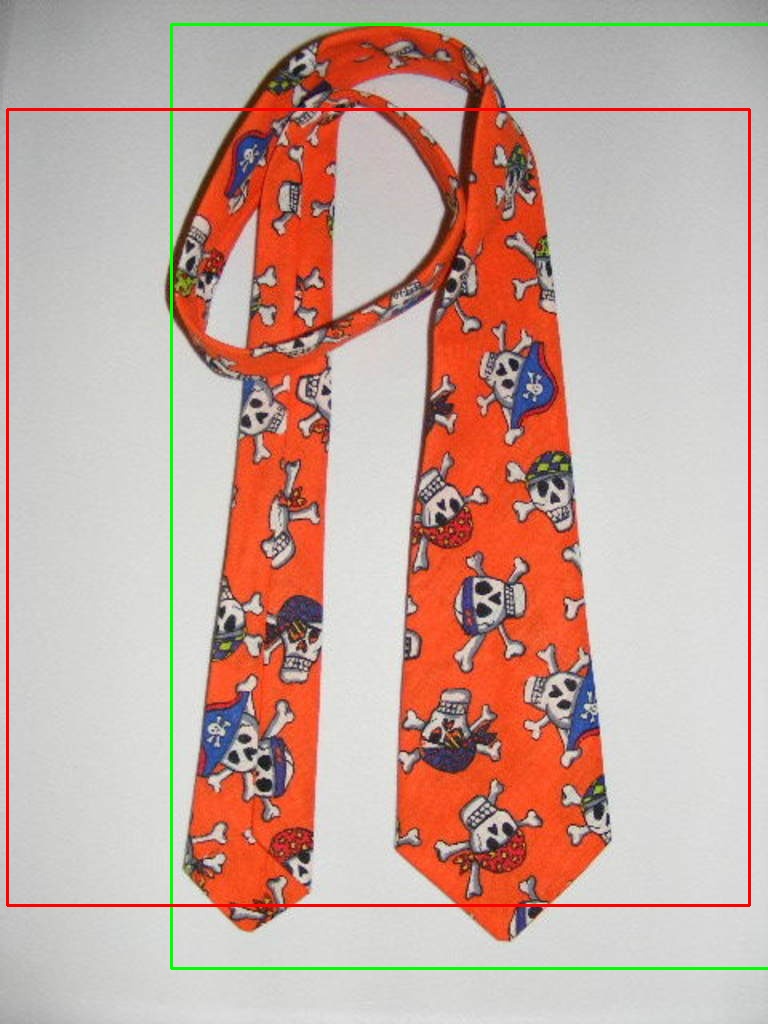} 
			\caption{Tie}
			\label{fig:accuracy_time_scratch_retinanet}
		\end{center}
	\end{subfigure}	
	\hfill
	\begin{subfigure}{.12\textwidth}
		\begin{center}
			\includegraphics[width=2.15cm,height=1.7cm]{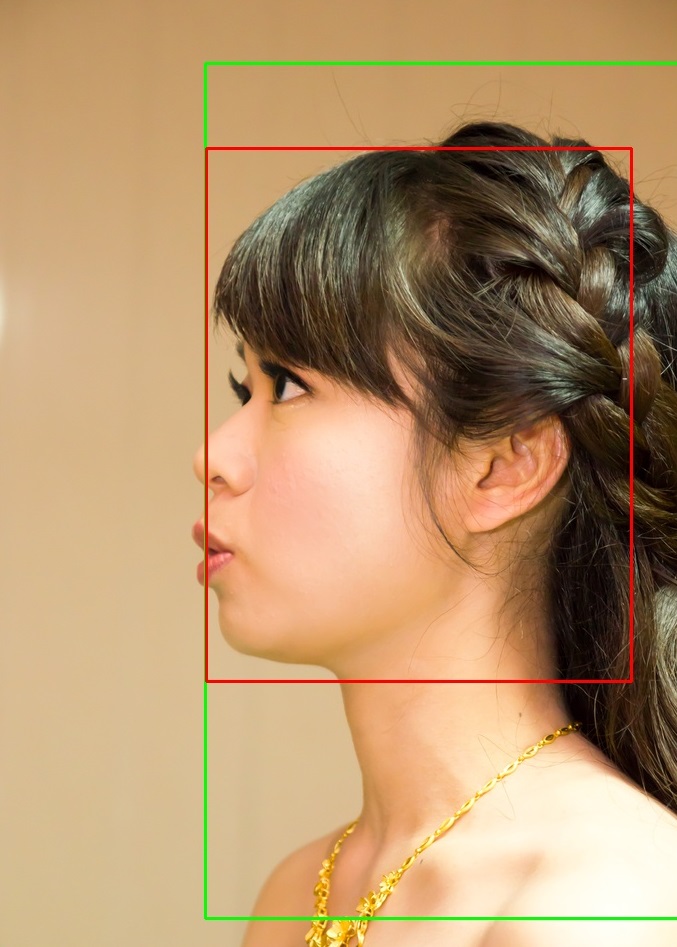} 
			\caption{Hairs}
			\label{fig:accuracy_time_scratch_retinanet}
		\end{center}
	\end{subfigure}	
	\hfill
	\begin{subfigure}{.12\textwidth}
		\begin{center}
			\includegraphics[width=2.15cm,height=1.7cm]{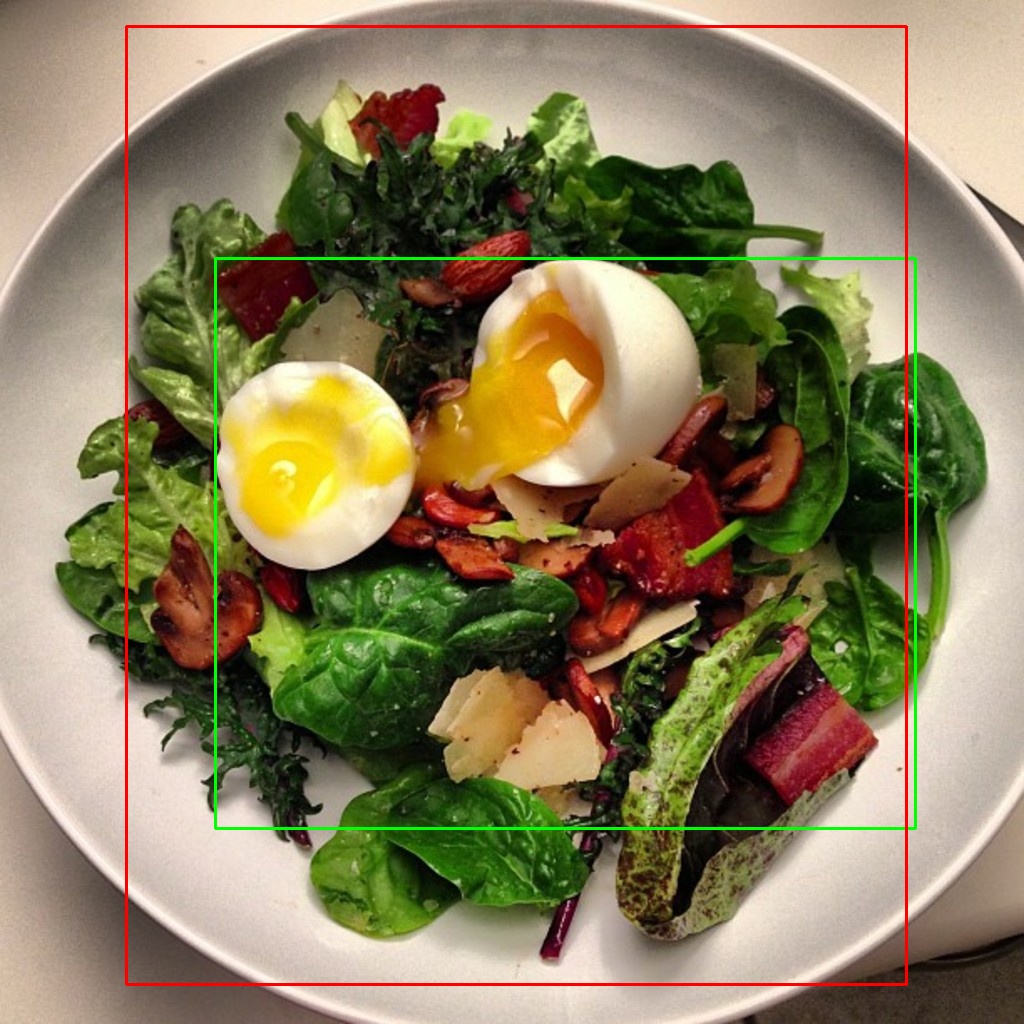} 
			\caption{Egg}
			\label{fig:accuracy_time_scratch_retinanet}
		\end{center}
	\end{subfigure}	
	\hfill
	\begin{subfigure}{.12\textwidth}
		\begin{center}
			\includegraphics[width=2.15cm,height=1.7cm]{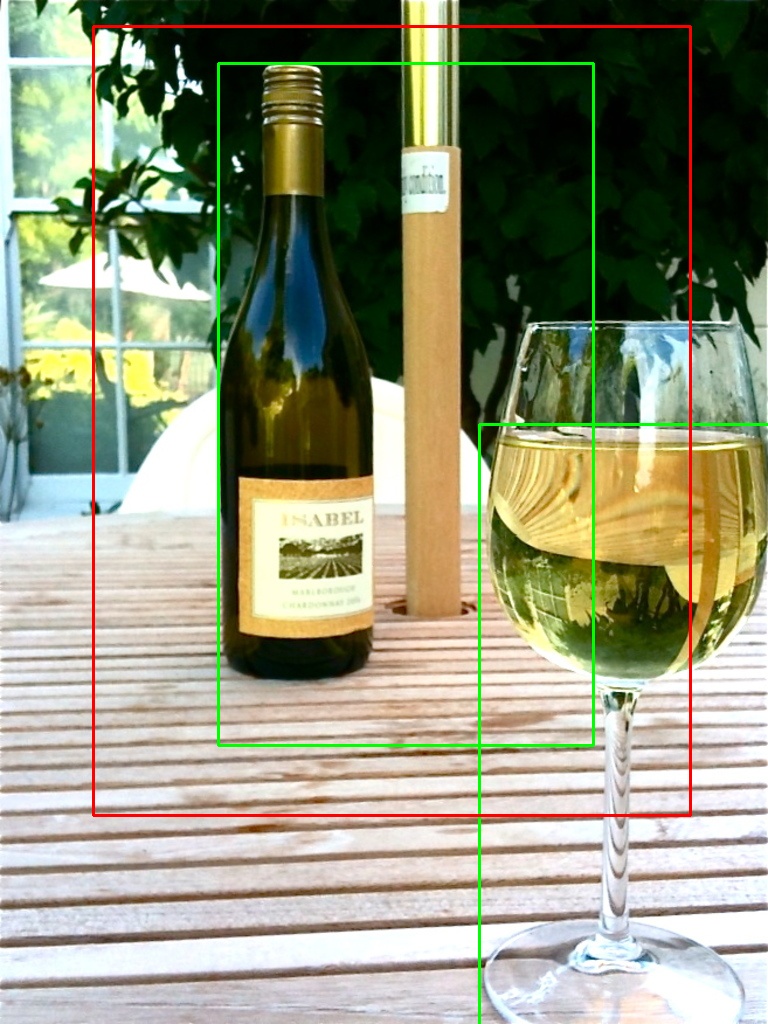} 
			\caption{Wine}
			\label{fig:accuracy_time_scratch_retinanet}
		\end{center}
	\end{subfigure}	
	
	\begin{subfigure}{.12\textwidth}
		\begin{center}
			\includegraphics[width=2cm,height=1.7cm]{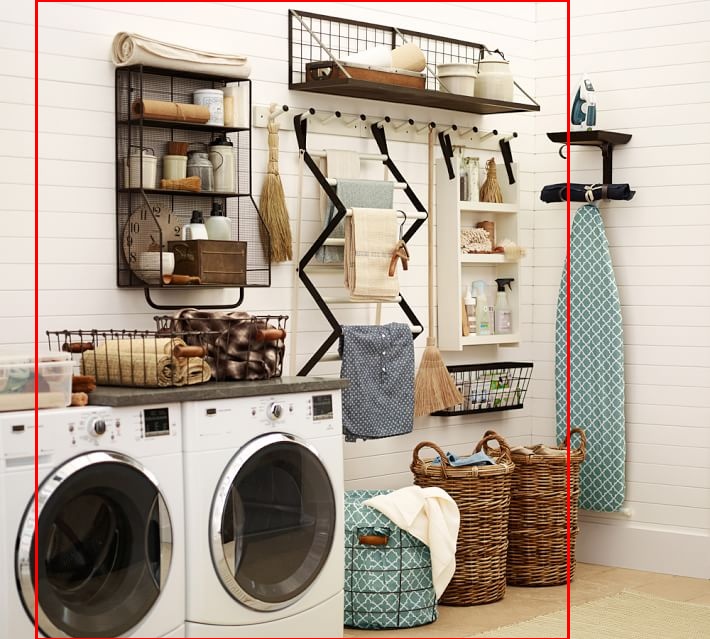} 
			\caption{IroningBoard}
			\label{fig:accuracy_time_scratch_yolo}
		\end{center}	
	\end{subfigure}%
	\hfill
	\begin{subfigure}{.12\textwidth}
		\begin{center}
			\includegraphics[width=2.15cm,height=1.7cm]{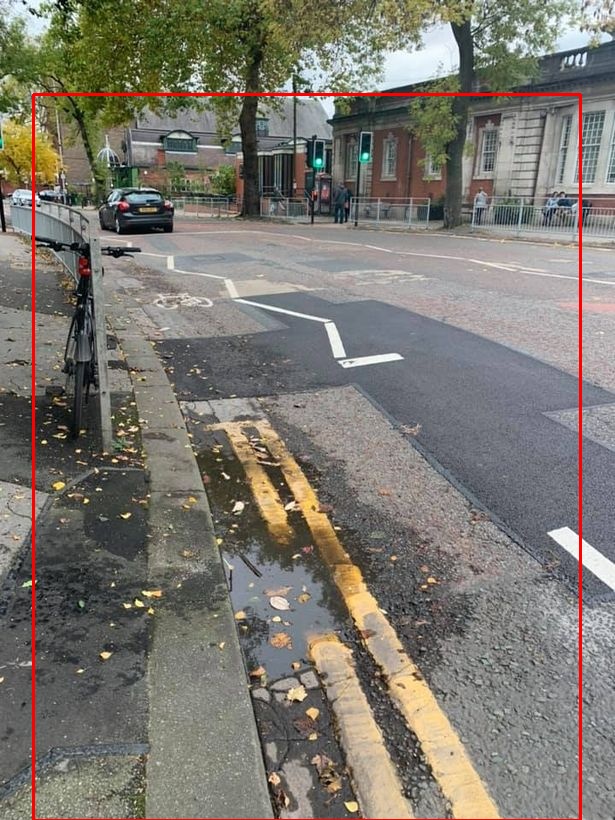} 
			\caption{Pothole}
			\label{fig:accuracy_time_scratch_ssd}
		\end{center}
	\end{subfigure}
	\hfill
	\begin{subfigure}{.12\textwidth}
		\begin{center}
			\includegraphics[width=2.15cm,height=1.73cm]{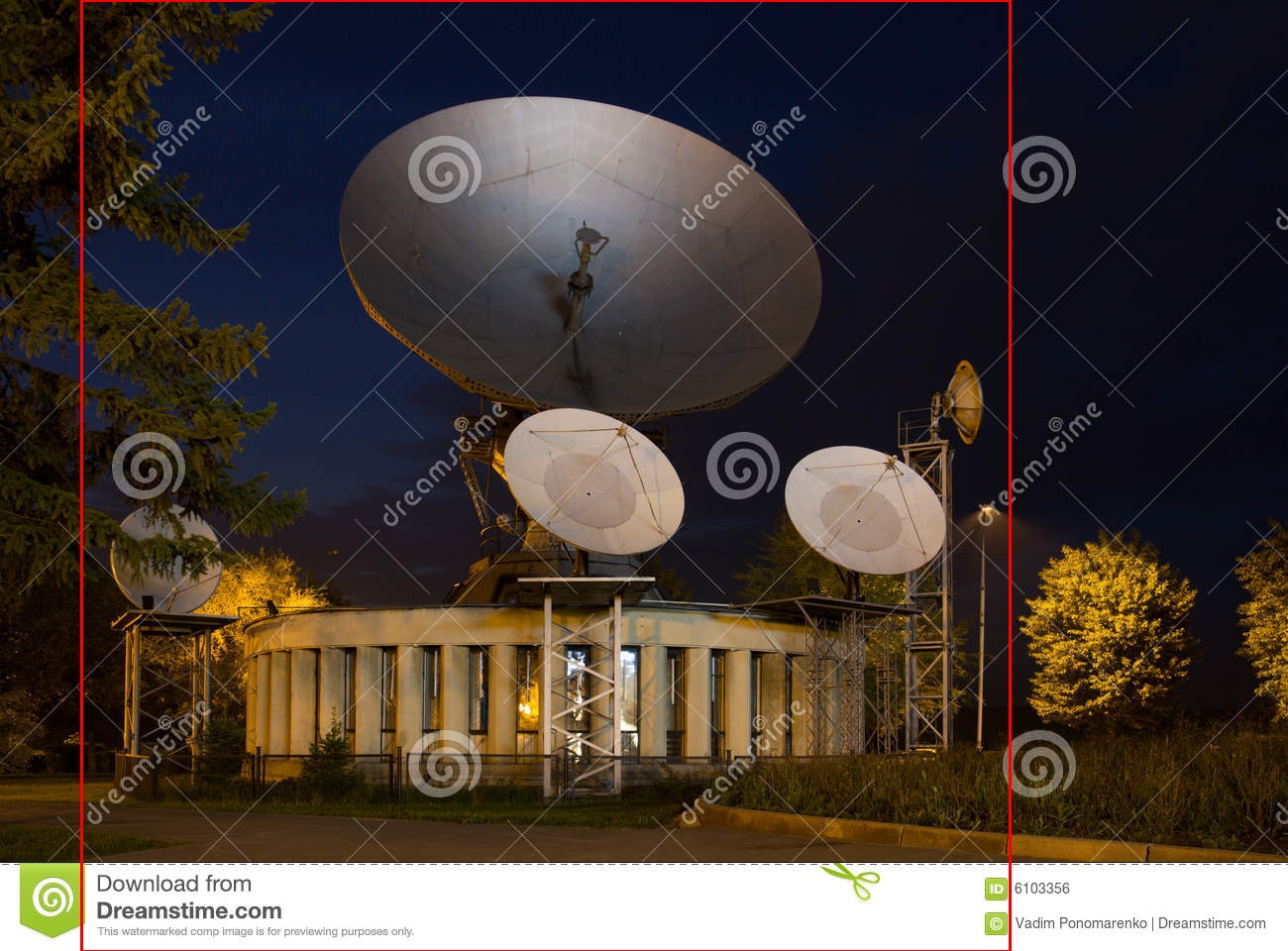} 
			\caption{SatelliteDish}
			\label{fig:accuracy_time_scratch_retinanet}
		\end{center}
	\end{subfigure}	
	\hfill
	\begin{subfigure}{.12\textwidth}
		\begin{center}
			\includegraphics[width=2.15cm,height=1.7cm]{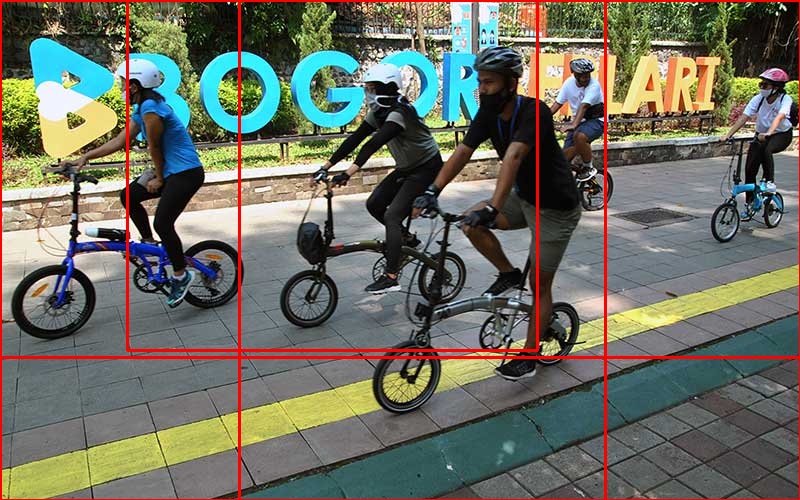} 
			\caption{Pedestrian}
			\label{fig:accuracy_time_scratch_retinanet}
		\end{center}
	\end{subfigure}	
	\hfill
	\begin{subfigure}{.12\textwidth}
		\begin{center}
			\includegraphics[width=2.13cm,height=1.7cm]{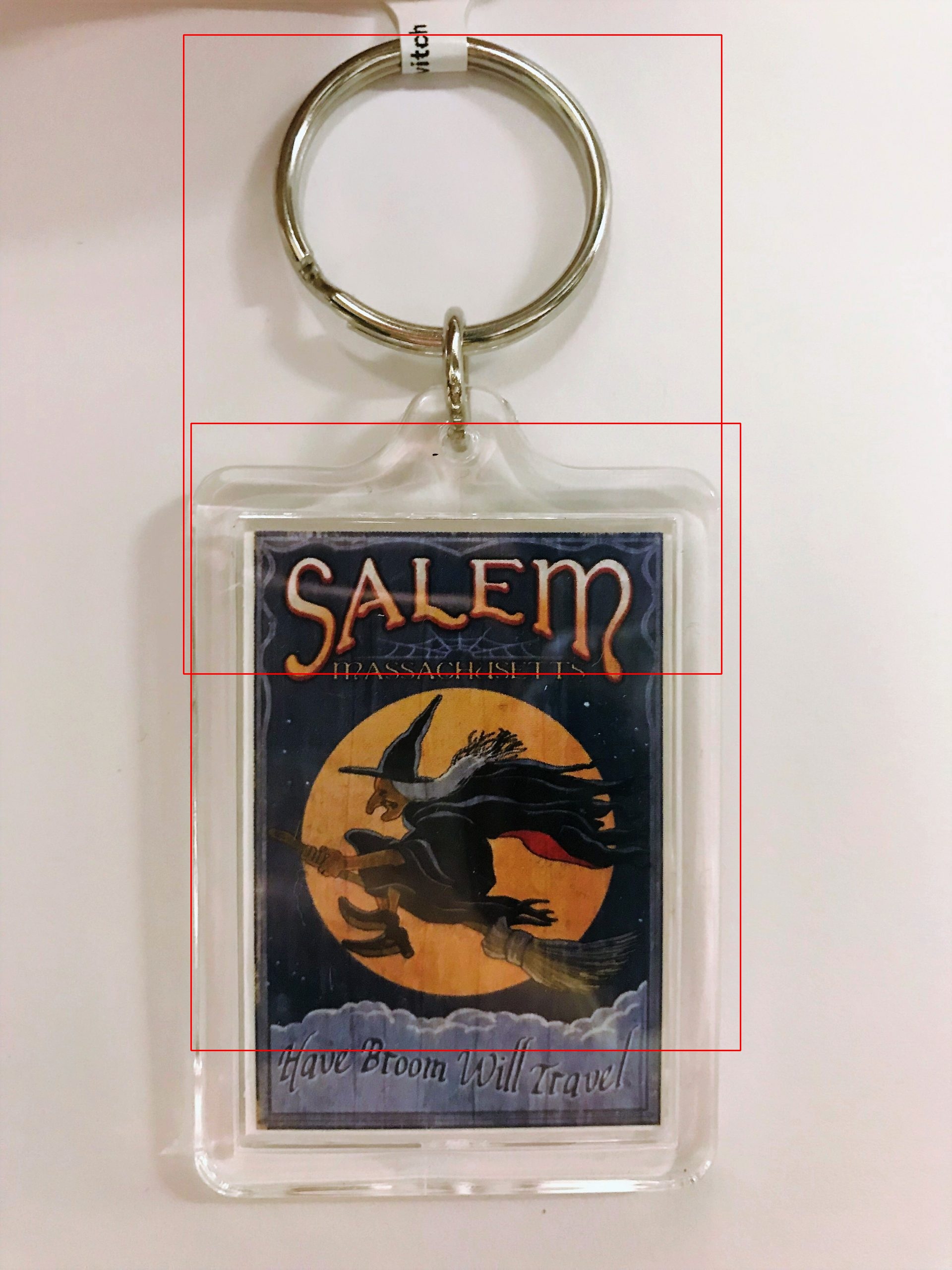} 
			\caption{Key}
			\label{fig:accuracy_time_scratch_retinanet}
		\end{center}
	\end{subfigure}	
	\hfill
	\begin{subfigure}{.12\textwidth}
		\begin{center}
			\includegraphics[width=2.15cm,height=1.7cm]{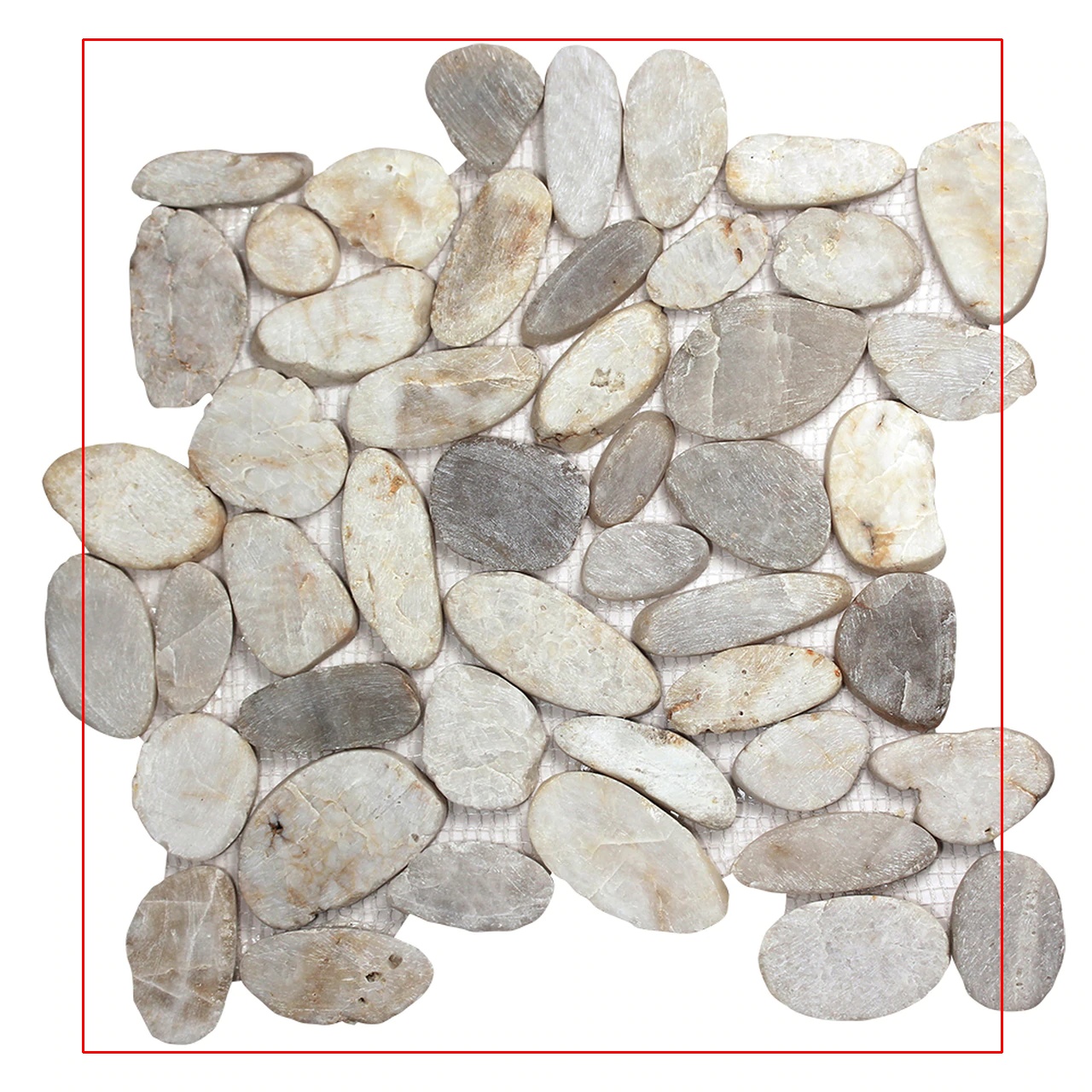} 
			\caption{Stone}
			\label{fig:accuracy_time_scratch_retinanet}
		\end{center}
	\end{subfigure}	
	\hfill
	\begin{subfigure}{.12\textwidth}
		\begin{center}
			\includegraphics[width=2.15cm,height=1.7cm]{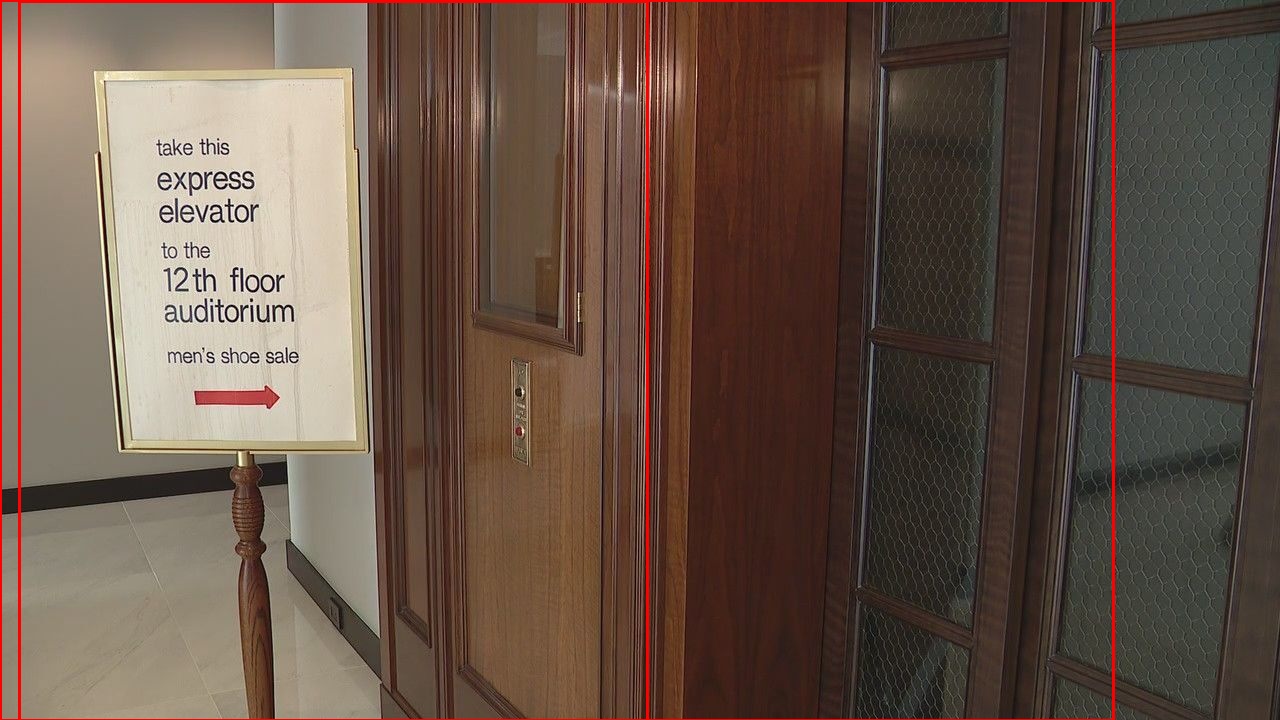} 
			\caption{Elevator}
			\label{fig:accuracy_time_scratch_retinanet}
		\end{center}
	\end{subfigure}	
	\hfill
	\begin{subfigure}{.12\textwidth}
		\begin{center}
			\includegraphics[width=2.15cm,height=1.7cm]{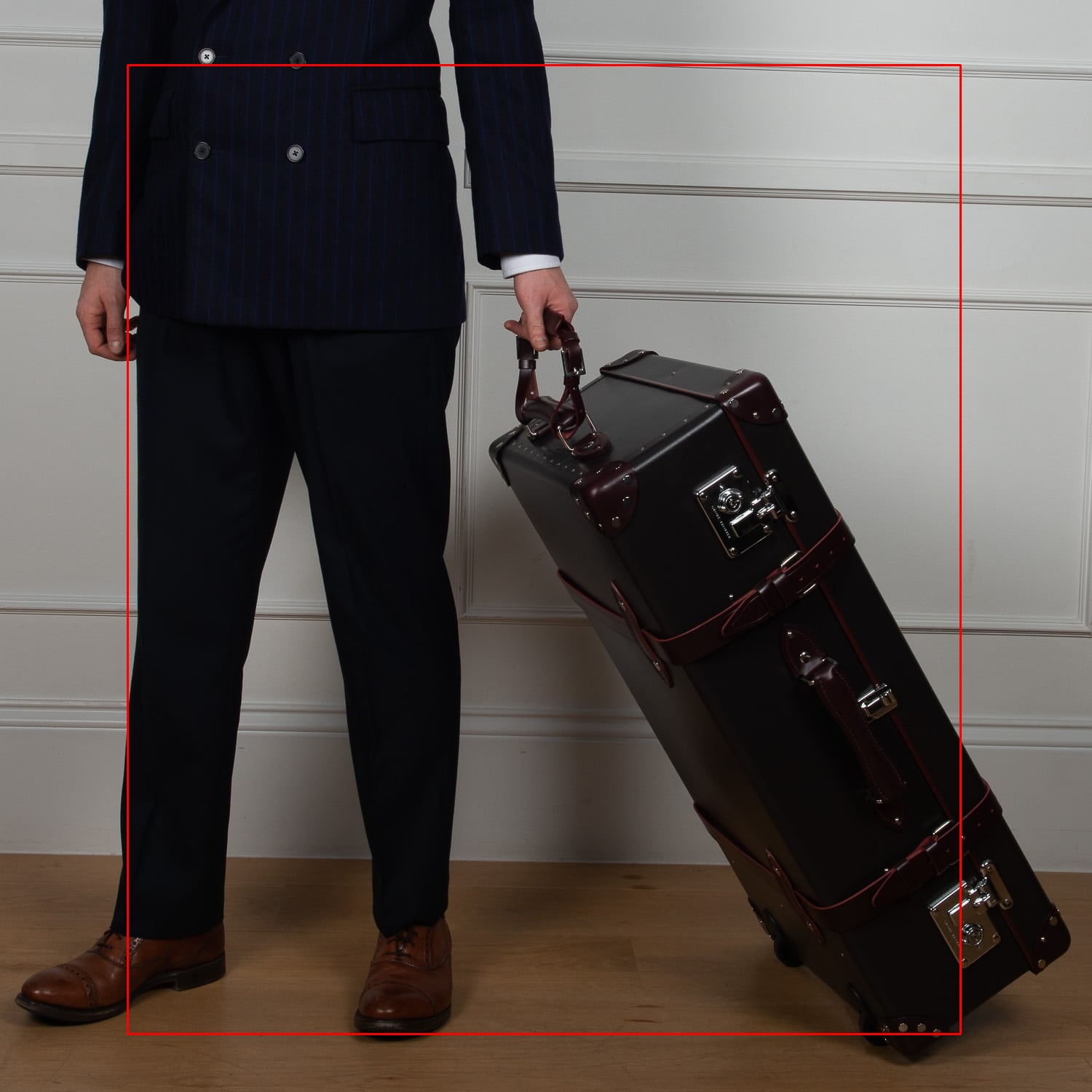} 
			\caption{Handle}
			\label{fig:accuracy_time_scratch_retinanet}
		\end{center}
	\end{subfigure}

	\caption{Examples of Incorrect location detections (label correct). 2nd row unseen classes does not have groundtruth.}
	\label{fig:example_incorrect_det}
\end{figure*}

\subsection{Qualitative Evaluation on Unseen Categories}
We show visual examples of our model detections in Figure \ref{fig:example_correct_det}. Examples of correct detections of our model on ``Unseen'' categories shown in red color and groundtruth (taken from OID) in green. Here Figure \ref{fig:example_correct_det} (a) -- (h) includes classes of ILSVRC, and Figure \ref{fig:example_correct_det} (i) -- (p) consist of additional unseen classes which not present in any object detection or image classification dataset to date. Correct detections of unseen concepts verify that UnseenNet can be trained on any class within a $10$ min of training. It also reduces the need to create large object detection datasets.

Some examples of incorrect detections are shown in Figure \ref{fig:example_incorrect_det}, where Figure \ref{fig:example_correct_det} (a) -- (h) includes classes of ILSVRC, and Figure \ref{fig:example_correct_det} (i) -- (p) consist of additional unseen classes that our model downloaded online and are not present in any object detection dataset to date. This demonstrates that if we have ``unseen'' classes less similar to ``seen'' classes, then UnseenNet could label them \textit{correctly} because of the training on classification data with \textit{incorrect} localization due to absence of detection data. 

\section{Conclusion and Future Work}
We presented an ``UnseenNet'' model that has the ability to construct a detector for any unseen concept without bounding boxes while training in a short time and providing competitive accuracy. We found that starting from a ``strong baseline detector'' trained on existing object detection datasets speed up the training rather than using only ImageNet \cite{ILSVRC15} pre-trained model to train unseen concepts. Moreover, in conjunction with semantic and visual similarity measures, classifier-detector conversion methods make our model more robust. Our evaluations demonstrate that \textit{UnseenNet} outperforms the baseline approaches in terms of training time for any unseen class and improves the mAP from 10\% to 30\% over existing detection based methods.


In the future, UnseenNet could be improved with more effective detectors and classifiers. Presently, we provided the size of images as bounding boxes for training weak baseline detector. This could be improved by background extraction or segmentation approaches and provide better annotations for training. Lastly, Strong and Weak baseline detectors could include large number of \textit{seen} classes to obtain more similar classes.


\bibliography{bibliography}
\bibliographystyle{icml2022}

%

\end{document}